\documentclass[final, journal, transmag]{support/IEEEtran}

\usepackage[markup=bfit,addedmarkup=colored,deletedmarkup=xout]{changes}

\usepackage{bm}
\usepackage{array}
\usepackage{makecell}
\usepackage{mathrsfs}
\usepackage{graphicx}
\usepackage{epstopdf}
\usepackage{amsmath,mathtools}
\usepackage{lipsum,adjustbox}
\usepackage{amsfonts}
\usepackage{amssymb}
\usepackage{calligra}
\usepackage{calrsfs}
\usepackage{cite}
\usepackage{float}
\usepackage{textcomp}
\usepackage{tikz}
\usepackage{subcaption}
\usetikzlibrary{shapes}
\usepackage{comment}
\usepackage[hidelinks]{hyperref}
\usepackage[hang,flushmargin]{footmisc}
\usepackage{tabularx,colortbl}
\usetikzlibrary{shapes.geometric, arrows, fit}
\usepackage{letltxmacro,xparse}

\graphicspath{{figures/}}

\definecolor{yellow}{RGB}{255,255,0}
\definecolor{red}{RGB}{255,91,51}
\definecolor{blue}{RGB}{8,191,223}
\definecolor{db}{RGB}{0,0,255}
\definecolor{green}{RGB}{54,203,30}
\definecolor{grey}{RGB}{170,170,170}
\definecolor{black}{RGB}{0,0,0}
\definecolor{orange}{RGB}{255,140,0}
\definecolor{yellow}{RGB}{225,249,27}

\tikzset{VertexStyle/.style = {shape = rectangle,fill = gray}}

\newcommand\blfootnote[1]{%
 \begingroup
 \renewcommand\thefootnote{}\footnote{#1}%
 \addtocounter{footnote}{-1}%
 \endgroup}

\def\footnoterule{\relax%
 \kern15pt
 \hbox to \columnwidth{\hfill\vrule width 1\columnwidth height 0.6pt\hfill}
 \kern4.6pt}

\LetLtxMacro\oldhref\href
\RenewDocumentCommand{\href}{o m m}{%
 \IfValueTF{#1}
 {\oldhref[#1]{#2}{\bfseries #3}}
 {\oldhref{#2}{\bfseries #3}}%
}

\DeclareMathOperator*{\argmin}{argmin}
\DeclareMathOperator*{\argmax}{argmax}

\newcommand{\apre}{{\textit{a priori }}}

\tikzstyle{arrow} = [thick,->,>=stealth]
\tikzstyle{circ} = [circle, minimum width=1cm, minimum height=1cm, text centered, draw=black, fill=blue!30]
\tikzstyle{circ_open} = [circle, minimum width=1cm, minimum height=1cm, text centered, draw=black, fill=red!60]
\tikzstyle{decision} = [diamond, minimum width=3cm, minimum height=1cm, text centered, draw=black, fill=orange!50,font=\bfseries]
\tikzstyle{process} = [rectangle, minimum width=6.0cm, minimum height=1cm, text centered, draw=black, fill=blue!60,font=\bfseries]
\tikzstyle{startstop} = [rectangle, rounded corners, minimum width=4.5cm, minimum height=1cm,text centered, draw=black, fill=green!50,font=\bfseries]

\usepackage{acronym}

    \acrodef{WVU}{West Virginia University}
    \acrodef{MAE}{Mechanical and Aerospace Engineering}

    \acrodef{IMU}{Inertial Measurement Unit}
    \acrodef{INS}{Inertial Navigation System}
    \acrodef{GPS}{Global Positioning System}
    \acrodef{GNSS}{Global Navigation Satellite System}
    \acrodef{LiDAR}{Light Detection And Ranging}
    
    \acrodef{PLL}{Phase Lock Loop}
    \acrodef{DLL}{Delay Lock Loop}
    \acrodef{IQ}{In-phase and Quadrature}
    \acrodef{SDR}{software defined radio}
    \acrodef{RINEX}{Receiver Independent Exchange Format}

    \acrodef{KF}{Kalman Filter}
    \acrodef{EKF}{Extended Kalman Filter}
    \acrodef{GMM}{Gaussian Mixture Model}
    \acrodef{NLLS}{Nonlinear Least Squares}
    \acrodef{SLAM}{Simultaneous Localization and Mapping}
    \acrodef{MAP}{maximum a posteriori}
    \acrodef{m-estimators}{Maximum Likelihood Estimators}
    \acrodef{IRLS}{iteratively re-weighted least squares}
    \acrodef{DCS}{dynamic covariance scaling}
    \acrodef{MM}{max-mixtures}
    \acrodef{RANSAC}{random sample consensus}
    \acrodef{RRR}{realizing, reversing, recovering}
    \acrodef{RAIM}{receiver autonomous integrity monitoring}
    \acrodef{BCE}{batch covariance estimation}

    \acrodef{MAP}{maximum a posteriori}
    \acrodef{MLE}{maximum likelihood estimate}
    \acrodef{PSD}{positive-semidefinite}

    \acrodef{RSOS}{Residual-Sum-of-Squares}

\begin{document}


\title{Enabling Robust State Estimation through Measurement Error Covariance Adaptation}







\author{Ryan~M.~Watson, Jason~N.~Gross, Clark~N.~Taylor,~and~Robert~C.~Leishman%
\thanks{R. Watson and J. Gross are with the Department of Mechanical and Aerospace Engineering, West Virginia University, 1306 Evansdale Drive, PO Box 6106, WV 26506-6106, United States. e-mail: rwatso12 (at) mix.wvu.edu \newline}
\thanks{C. Taylor is with the Department of Electrical and Computer Engineering, Air Force Institute of Technology, 2950 Hobson Way, WPAFB, OH 45433-7765, United States \newline}
\thanks{R. Leishman is with the Autonomy and Navigation Technology Center, Air Force Institute of Technology, 2950 Hobson Way, WPAFB, OH 45433-7765, United States.}}

\maketitle
\markboth{Submitted to IEEE Transactions on Aerospace and Electronic Systems}%
{Watson \MakeLowercase{\textit{et al.}}: Enabling Robust State Estimation through Measurement Error Covariance Adaptation}

\begin{abstract}

Accurate platform localization is an integral component of most robotic systems. As these robotic systems become more ubiquitous, it is necessary to develop robust state estimation algorithms that are able to withstand novel and non-cooperative environments. When dealing with novel and non-cooperative environments, little is known {\textit{a priori}} about the measurement error uncertainty, thus, there is a requirement that the uncertainty models of the localization algorithm be adaptive. Within this paper, we propose \replaced{the batch covariance estimation technique, which}{ one such technique that} enables robust state estimation through the iterative adaptation of the measurement uncertainty model. The adaptation of the measurement uncertainty model is granted through non-parametric clustering of the residuals, which enables the characterization of the measurement uncertainty via a Gaussian mixture model. The provided Gaussian mixture model can be utilized within any non-linear least squares optimization algorithm by approximately characterizing each observation with the sufficient statistics of the assigned cluster (i.e., each observation's uncertainty model is updated based upon the assignment provided by the non-parametric clustering algorithm). The proposed algorithm is verified on several GNSS collected data sets, where it is shown that the proposed technique exhibits some advantages when compared to other robust estimation techniques when confronted with degraded data quality.

\end{abstract}

\section{Introduction}\label{sec:introduction}

The applicability of robotic platforms to an ever increasing number of applications (e.g., disaster recovery \cite{murphy2014disaster}, scientific investigations \cite{wettergreen2014science}, health care \cite{riek2017healthcare}) has been realized in recent years. A core component that enables the operation of these platforms is the ability to localize (i.e., estimate the position states within a given coordinate frame) given {\textit{a priori}} information and a set of measurements. Thus, a considerable amount of research has been afforded to the problem of accurate and robust localization of a robotic platform.

To facilitate the localization of a platform, a state estimation \cite{simon2006optimal} scheme must be implemented. These state estimation frameworks can take two forms: either a batch estimator, or an incremental estimator. The batch variant of state estimation is generally facilitated through the utilization of a \ac{NLLS} algorithm (e.g., line-search method \cite{de2004nonlinear}, or a trust-region method \cite{yuan2000review}). On the other hand, incremental state estimators are usually implemented as a variant of the Kalman filter \cite{kalman} (e.g., the extended Kalman filter \cite{bertsekas1996incremental}, or the unscented Kalman filter \cite{julier1997new}).

The cost function underlying all of the state estimation techniques mentioned above is the $l^2$-norm of the estimation errors.  This cost function enables accurate and efficient estimation when the assumed models precisely characterize the provided observations. However, when the utilized model does not accurately characterize the provided measurements, the estimation framework can provide an arbitrarily poor state estimate. Specifically, as discussed in \cite{graham2015robust}, the $l^2$-norm cost function has an asymptotic breakdown of zero (i.e., if any single observation deviates from the utilized model, the estimated solution can be biased by an arbitrarily large quantity \cite{hampel1968contribution}).

To combat the poor breakdown properties of the $l^2$-norm cost function, several robust estimation frameworks have been proposed. These frameworks can be broadly partitioned into two categories: data weighting methods, and data exclusion methods. The data weighting methods work by iteratively calculating a measurement weighting vector such that observations which most substantially deviated form the utilized model have a reduced influence on the estimated result. Several commonly utilized data weight techniques include: robust \ac{m-estimators} \cite{huberBook}, switchable constraints \cite{sunderhauf2012robust}, \ac{DCS} \cite{DCS}, and max mixtures \cite{maxmix}. On the other hand, the data exclusion methods work by finding a trusted subset of the provided observations. Several commonly utilized data exclusion techniques include: \ac{RANSAC} \cite{fischler1981random}, \ac{RRR} \cite{latif2012realizing}, $l^1$ relaxation \cite{carlone2014selecting}, and \ac{RAIM}\cite{zhai2018fault}.

\blfootnote{\\ All developed software and data utilized within this study is publicly available at \href{https://bit.ly/2V4QuB0}{https://bit.ly/2V4QuB0}. }

Both estimation paradigms have certain undesirable properties. For example, in many applications, it is undesirable to completely remove observations that do not adhere to the defined model (e.g., the utilization of \ac{GNSS} in an urban environment where it may not be possible to estimate the desired states if observations are removed). Instead, it can be more informative to accurately characterize the measurement uncertainty model and accordingly reduce the observations influence on the estimate. This concept of data uncertainty model characterization naturally conforms to the data weighting class of robust techniques (as discussed more thoroughly in section \ref{subsec:robust_est}). However, the currently utilized weighting techniques have the undesirable property that they assume an accurate characterization of the true measurement uncertainty model can be provided {\textit{a priori}}, which may not be a valid assumption if the robotic platform is operating in a novel environment.

Within this paper, we expand upon our prior work in \cite{watsonbatch} on a novel robust data weighting method, which relaxes the assumption that an accurate {\textit{a priori}} measurement uncertainty characterization can be provided. This assumption is relaxed by iteratively estimating a \ac{GMM} based on the state estimation residuals that characterizes the measurement uncertainty model. This concept was also studied, following the work of \cite{watsonbatch}, within \cite{pfeifer2018expectation} where an expectation-maximization (EM) clustering algorithm was implemented to characterize the measurement uncertainty model. The approach provided in this paper extends upon the work in \cite{watsonbatch} on two fronts: first, a computationally efficient variational approach \cite{kurihara2007accelerated,steinberg2013unsupervised} is utilized for \ac{GMM} model fitting as opposed to the collapsed Gibbs sampling \cite{liu1994collapsed} as previously utilized within \cite{watsonbatch}, and secondly, the proposed approach is thoroughly verified on a diverse set of kinematically collected GNSS observations.

To facilitate a discussion of the proposed robust estimation technique, the remainder of this paper proceeds in the following manner. First, in Section \ref{sec:state_estimation}, an succinct overview of \ac{NLLS} is provided, with a specific emphasis placed on robust estimation. In Section \ref{sec:technical_approach}, a novel robust estimation methodology, which enables adaptive observation de-weighting though the estimation of a multimodal measurement error covariance, is proposed. In Sections \ref{sec:experimental_setup} and \ref{sec:results}, the proposed approach is validated with multiple kinematic \ac{GNSS} data-sets. Finally, the paper concludes with final remarks and proposed future research efforts.

\section{State Estimation}\label{sec:state_estimation}

The traditional state estimation problem is concerned with finding the set of states, $X$, that, in some sense, best describe the set of observations, $Y$. An intuitive, and commonly \replaced{utilized}{ utilize} method to quantify the level of fit between the set of states and the set of observations is the evaluation of the posterior distribution, $\operatorname{p}(X \ | \ Y)$. When the state estimation problem is cast in this light, the desired set of states can be calculated through the maximization of the posterior distribution, as provided in Eq. \ref{eq:map_cost}, which is generally referred to as the \ac{MAP} state estimate \cite{simon2006optimal}.

\begin{equation}
 \hat{X} = \argmax_X \ \operatorname{p}(X \ | \ Y)
 \label{eq:map_cost}
\end{equation}

A commonly utilized way to efficiently formulate the \ac{MAP} estimation problem is through the application of the factor graph \cite{ogFG}. The factor graph enables efficient estimation through the factorization of the posterior distribution into the product of local functions (i.e., functions operating on a reduced domain when compared to the posterior distribution)

\begin{equation}
 \operatorname{p}(X \ | \ Y) \propto \prod_{n=1}^{N} \psi_i(A_n,B_n),
 \label{eq:fg_factorization}
\end{equation}

\noindent where, $A_n \subseteq \{ X_1, X_2 \ldots, X_n\}$, and $B_n \subseteq \{ Y_1, Y_2 \ldots Y_m \}$. It should be noted that the above expression could be an equality through multiplication by a normalization constant.

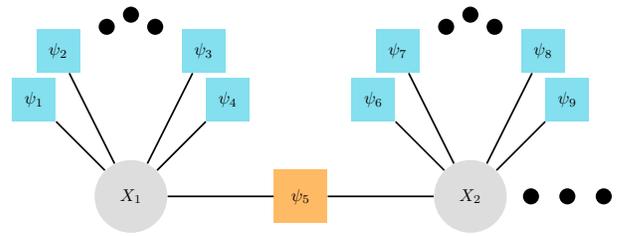
\begin{figure}
 \centering
 \begin{adjustbox}{width=0.9\linewidth}
  \begin{tikzpicture}

   \node[shape=circle,fill=grey!40, minimum size=1.5cm] (B) at (4,0) {$X_{1}$};
   \node[shape=circle,fill=grey!40, minimum size=1.5cm] (C) at (11,0) {$X_{2}$};

   \node[shape=circle,fill=black,minimum size=0.05cm] (x) at(12.25,0) {};
   \node[shape=circle,fill=black,minimum size=0.05cm] (y) at(13.0,0) {};
   \node[shape=circle,fill=black,minimum size=0.05cm] (z) at(13.75,0) {};

   \node[shape=rectangle,fill=blue!50, minimum size=0.9cm] (H) at (2,2) {$\psi_{1}$};
   \node[shape=rectangle,fill=blue!50, minimum size=0.9cm] (I) at (2.5,3) {$\psi_{2}$};
   \node[shape=circle,fill=black,minimum size=0.05cm] (a) at(3.5,3.5) {};
   \node[shape=circle,fill=black,minimum size=0.05cm] (b) at(4,3.75) {};
   \node[shape=circle,fill=black,minimum size=0.05cm] (c) at(4.5,3.5) {};
   \node[shape=rectangle,fill=blue!50, minimum size=0.9cm] (J) at (5.5,3) {$\psi_{3}$} ;
   \node[shape=rectangle,fill=blue!50, minimum size=0.9cm] (K) at (6,2) {$\psi_{4}$};

   \node[shape=rectangle,fill=orange!60, minimum size=1.1cm] (Q) at (7.5,0) {$\psi_{5}$};

   \node[shape=rectangle,fill=blue!50, minimum size=0.9cm] (L) at (9,2) {$\psi_{6}$};
   \node[shape=rectangle,fill=blue!50, minimum size=0.9cm] (M) at (9.5,3) {$\psi_{7}$};
   \node[shape=circle,fill=black,minimum size=0.05cm] (a) at(10.5,3.5) {};
   \node[shape=circle,fill=black,minimum size=0.05cm] (b) at(11,3.75) {};
   \node[shape=circle,fill=black,minimum size=0.05cm] (c) at(11.5,3.5) {};
   \node[shape=rectangle,fill=blue!50, minimum size=0.9cm] (N) at (12.5,3) {$\psi_{8}$};
   \node[shape=rectangle,fill=blue!50, minimum size=0.9cm] (O) at (13,2) {$\psi_{9}$};

   \path [-,line width=1pt] (B) edge node {} (H);
   \path [-,line width=1pt] (B) edge node {} (I);
   \path [-,line width=1pt] (B) edge node {} (J);
   \path [-,line width=1pt] (B) edge node {} (K);

   \path [-,line width=1pt] (B) edge node {} (Q);
   \path [-,line width=1pt] (Q) edge node {} (C);

   \path [-,line width=1pt] (C) edge node {} (L);
   \path [-,line width=1pt] (C) edge node {} (M);
   \path [-,line width=1pt] (C) edge node {} (N);
   \path [-,line width=1pt] (C) edge node {} (O);

  \end{tikzpicture}
 \end{adjustbox}
 \caption{Example  posterior distribution factorization with a factor graph for a GNSS example, where each $X_n$ represents a set of states to be estimated, and each $\psi_n$ represents a constraint on the associated (i.e., connected with an edge) state vectors.}
 \label{fig:factor_graph}
\end{figure}

More formally, the factor graph is a bipartite graph with state nodes, $X$, and constraint nodes, $\psi$. Additionally, edges only exist between dependent state and constraint nodes \cite{dellaert2017factor}. This factorization is visually depicted in Fig. \ref{fig:factor_graph}, for a simple \ac{GNSS} example. Within Fig. \ref{fig:factor_graph}, each $X$ represents a state vector to be estimated, and $\psi$ represents the constraints on a subset of the estimated states (e.g., $\psi_n$ could represent a \ac{GNSS} pseudorange observation constraint on state vector $X_n$,  or an inertial measurement linking two states together over time).

Utilizing the factorization specified in Eq. \ref{eq:fg_factorization} for the posterior distribution, the modified state estimation problem takes the form presented in Eq. \ref{eq:fg_optimization}.

\begin{equation}
 \hat{X} = \argmax_X \prod_{n=1}^{N} \psi_n(A_n,B_n)
 \label{eq:fg_optimization}
\end{equation}

In practice, this state estimation problem is simplified through the assumption that each factor adheres to a Gaussian uncertainty model (i.e., $\psi_n(A_n,B_n) \sim \mathcal{N}(\mu_n,\Lambda_n)$) \cite{dellaert2017factor}. When this assumption is enforced, the \ac{MAP} estimation problem is equivalent to the minimization of the weighted sum of squared residuals \cite{dellaert2017factor}, as provided in Eq. \ref{eq:nlls_cost}

\begin{equation}
 \hat{X} = \argmin_X \sum_{n=1}^{N} \lvert \lvert \ r_n(X) \ \rvert \rvert_{\Lambda_n} \quad \text{s.t.} \quad r_n(X) \triangleq y_n - h_n(X),
 \label{eq:nlls_cost}
\end{equation}

\noindent where, within this paper, the $\lvert \lvert \ * \ \rvert \rvert$ function is used to define the $l^2\text{-norm}$, $h_n$ is a mapping from the state space to the observation space, and $\Lambda_n$ is the provided \added{symmetric, \ac{PSD}} covariance matrix \added{, which is assumed to be diagonal}.

With a cost function as presented in Eq. \ref{eq:nlls_cost}, a \ac{MAP} state estimate can be calculated using any \ac{NLLS} algorithm. To enable the selection of a \ac{NLLS} algorithm, it is informative to coarsely partition the class of algorithms into two categories: line search techniques, and trust region techniques. The line search techniques operate by iteratively updating the previous state estimate by calculating a step direction $\Delta_{x_k}$ and a step magnitude, $\alpha_k$,  (i.e., $\hat{X}_{k+1} = \hat{X}_{k} + \alpha_k \Delta_{x_k}$), where the specific calculation of $\alpha_k$, and $\Delta_{x_k}$ differs for varying algorithms (e.g., see the Gauss-Newton implementation \cite{madsen1999methods} for one such example). The trust region based \ac{NLLS} methods incorporate an additional constraint that the state update at the current iteration must fall within a ball of the current estimate (i.e.,  \replaced{$|| \alpha_k \Delta_{x_k} || < r$}{$|| \alpha_k \Delta_{x_k} ||_2 < r$}, where $r$ is the trust region radius). For trust region based methods, the Levenberg Marquardt \cite{more1978levenberg} -- which is the \ac{NLLS} algorithm utilized within this study -- and the Dog-leg \cite{dogleg} approaches are commonly utilized in practice.

\subsection{Robust State Estimation Through Covariance Adaptation} \label{subsec:robust_est}

One of the major disadvantages of utilizing the $l^2\text{-norm}$ cost function is the undesirable asymptotic breakdown property \cite{hampel1968contribution}. Specifically, any estimator solely utilizing an $l^2\text{-norm}$ cost function will have an asymptotic breakdown point of zero. As discussed in \cite{graham2015robust}, this property can be intuitively understood by letting any arbitrary observation, $y_n$, significantly deviate from the model (\replaced{i.e.,}{ i..e.,} $\lvert \lvert y_n - h_n(X) \rvert \rvert_{\Lambda_n} \rightarrow \infty$), which, in turn,  will \added{arbitrarily bias} the provided state estimate \replaced{(i.e., $\lvert \lvert X \rvert \rvert \rightarrow \infty$).}{ $\hat{X} = \sum_{n=1}^N r_n(X)^T \Lambda_n^{-1} (y_n - h_n(X)) \rightarrow \infty.$}

The undesirable breakdown property of the $l^2\text{-norm}$ cost function has spurred significant research interest in the direction of robust state estimation \cite{huberBook, sunderhauf2012robust, DCS, maxmix}. Within this paper, several of the major advances within the field will be discussed from the point-of-view of covariance adaptation; however, it should be noted that this is not an exhaustive compilation of all robust state estimation techniques. Rather, the purpose of this discussion is to provide a concise overview of the recent advances within the field, with specific emphasis placed on robust state estimation contributions now commonly adopted within the robotics community.

\subsubsection{M-Estimators} \label{subsubsec:m-est}

\begin{figure}
 \centering
 \includegraphics[width=.9\linewidth]{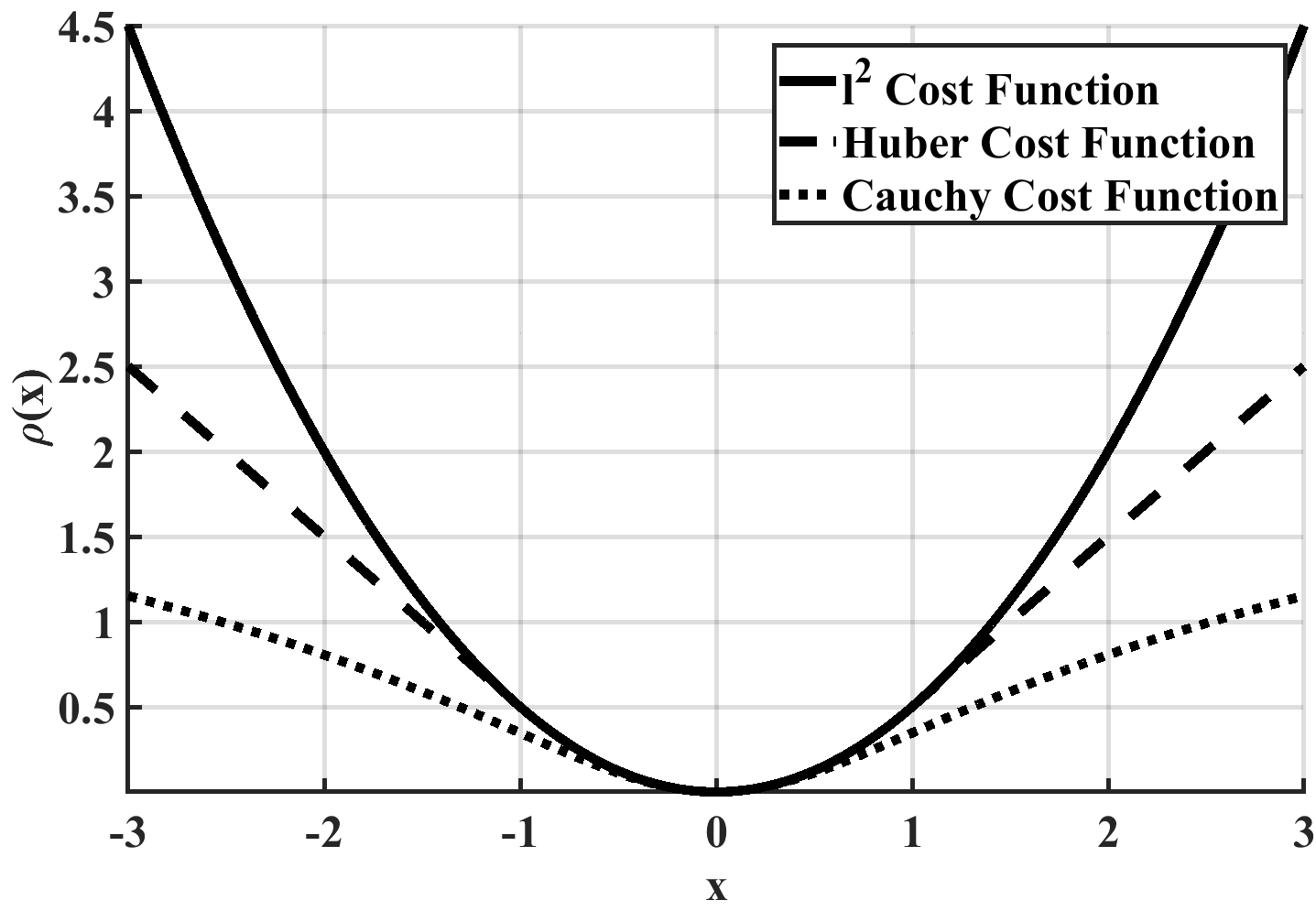}
 \caption{The cost function response for two commonly utilized robust \ac{m-estimators}. Robust in this sense implies that the cost functions increases more slowly than the $l^2$-norm for large values of $x$. The specific form of the \ac{m-estimators} is presented in Table \ref{table:m-estimators}.}
 \label{fig:m_estimators}
\end{figure}

One of the initial frameworks developed to enable robust state estimation is through a class of \ac{m-estimators}, as introduced by Huber \cite{huber64, huberBook}, that are less sensitive to erroneous observations\footnote{Erroneous observations are simply observations that do not adhere to the defined observation model.} than the $l^2$-norm. Specifically, this framework aims to replace the $l^2$-norm cost function, with a modified cost function that is more robust (i.e., increases more slowly when compared to the $l^2$-norm cost function, as depicted in Fig. \ref{fig:m_estimators}). This class of modified cost functions, generically, take a form as presented in Eq. \ref{eq:m-estimator}, where $\rho(*)$ is the modified cost function.

\begin{align}
 \hat{X} & = \argmin_X \sum_{n=1}^N \rho \big( \ \lvert \lvert r_n(X) \rvert \rvert_{\Lambda_n} \ \big) \nonumber \\&= \argmin_X \sum_{n=1}^{N} \rho( e_n ) \quad \text{s.t.} \quad e_n \triangleq \lvert \lvert r_n(X) \rvert \rvert_{\Lambda_n}
 \label{eq:m-estimator}
\end{align}

While numerous robust cost functions exist, we focus on two commonly utilized functions: the Huber and Cauchy cost functions.  (For additional \ac{m-estimators}, the reader is referred to \cite{mactavish2015all}.)  Associated with each robust cost function is an influence and weighting function that can be used to enable these cost functions to be solved with the same \ac{NLLS} approaches used for a quadratic cost function.  A summary of the cost, influence and weight function is shown in Table \ref{table:m-estimators}.


\begin{table}[htb!]
 \renewcommand{\arraystretch}{1.5}
 \centering
 \caption{Functions utilized to implement the $l^2$-norm, Cauchy, and  Huber cost functions, where, where $\rho(x)$ is the robust cost function, $\psi(x) \triangleq \frac{\partial \rho(x)}{\partial x}$ is the influence function, $w(x) \triangleq \frac{\psi(x)}{x}$ is the weighting function, and $k$ is the user defined kernel width.   }
 \resizebox{\columnwidth}{!}{
  \begin{tabular}{|| c | c | c | c ||}
  \hline
  M-Estimator                 & $\rho(x)$                             & $\psi(x)$                             & $w(x)$                                \\ [0.5ex]
  \hline\hline
  $l^2\text{-norm}$ & $x^2 \mathbin{/} 2$                   & $x$                                   & $1$                                   \\
  \hline
  Huber $\left\{ \begin{array}{@{}ll@{}} \text{if} \ \ |x| \leq k \\ \text{if} \ \ |x| > k \end{array}\right.$
  & $\left\{ \begin{array}{@{}ll@{}} \frac{1}{2}x^2 \\ k(|x|-\frac{k}{2}) \end{array}\right.$  &
  $\left\{ \begin{array}{@{}ll@{}} x \\ k\operatorname{sgn}(x) \end{array}\right.$
  & $\left\{ \begin{array}{@{}ll@{}} 1 \\ k\mathbin{/}|x| \end{array}\right.$ \\
  \hline
  Cauchy                      & $\frac{k^2}{2}log(1+\frac{x^2}{k^2})$ & $x \mathbin{/} \big[ 1+x^2/k^2 \big]$ & $1 \mathbin{/} \big[ 1+x^2/k^2 \big]$ \\ [1ex]
  \hline
  \end{tabular}%
 }
 \label{table:m-estimators}
\end{table}

As discussed at the outset of this section, our objective is to view robust state estimation as a type  of covariance adaptation. To view \ac{m-estimators} as a type of covariance adaptation, we will begin by take the gradient of the modified cost function (see Eq. \ref{eq:m-estimator}), as presented in Eq. \ref{eq:m-est_grad}.

\begin{align}
 \frac{\partial J}{\partial X} & = \sum_{n=1}^{N} \frac{\partial \rho}{\partial e_n} \frac{\partial e_n}{\partial r_n} \frac{\partial r_n}{\partial X} \nonumber \\&=  r_n(X)^{T} \bigg[ \frac{1}{e_n} \left. \frac{\partial \rho}{\partial e} \right|_{e_n(X)} \Lambda_n^{-1} \bigg] \frac{\partial r_n}{\partial X}.
 \label{eq:m-est_grad}
\end{align}

The expression for the gradient of the modified cost function can be further reduced by noting that $\frac{1}{e_n} \frac{\partial \rho}{\partial e_n}$, is equivalent to the m-estimator weighting function, by definition. This simplifies the gradient expression for the \ac{m-estimators} cost function as presented in Eq. \ref{eq:m-est_grad_final}, which is equivalent to the traditional $l^2$-norm cost function, but with an adaptive covariance. The modified covariance, $\hat{\Lambda}_n$, is provided in Eq. \ref{eq:m-est_cov_scale}, where the specific weighting function is dependent upon the utilized m-estimator.

\begin{equation}
 \frac{\partial J}{\partial X} = r_n(X)^T \bigg[ w\big(e_n\big) \Lambda_n^{-1} \bigg] \frac{\partial r_n}{\partial X},
 \label{eq:m-est_grad_final}
\end{equation}

\begin{equation}
 \hat{\Lambda}_n = \big[ w(e_n) \Lambda_j^{-1} \big]^{-1} = \frac{1}{w(e_n)} \Lambda_j,
 \label{eq:m-est_cov_scale}
\end{equation}

\noindent From this discussion, it should be noted that the robust state estimation problem when \ac{m-estimators} are utilized is equivalent to an \ac{IRLS} problem \cite{holland1977robust,bosse2016robust}. In \ac{IRLS}, the effect of an erroneous observation is minimized by covariance adaptation based on the previous optimization iteration's residuals, as presented in Eq. \ref{eq:irls}. The equivalence between \ac{m-estimators} and \ac{IRLS} has been previously noted in the literature \cite{zhang1997parameter,barfoot2017state}.

\begin{equation}
 \hat{X} = \argmin_X \sum_{n=1}^{N} w_n(e_{n}) \ \lvert \lvert \ r_n(X) \ \rvert \rvert_{\Lambda_n}
 \label{eq:irls}
\end{equation}

\subsubsection{Switch Constraints}

A more recent advancement within the robotics community, switch constraints (as introduced in \cite{switchCon}) is a specific realization of a lifted state estimation scheme \cite{zach2017iterated}. Switch constraints augment the optimization problem by concurrently solving for the desired states and a set of measurement weighting values, as provided in Eq. \ref{eq:switch_con}

\begin{equation}
\hat{X},\hat{S}  = \argmin_{X,S} \sum_{n=1}^N \bigg[ \lvert \lvert \ \psi(s_n)r_n(X) \ \rvert \rvert_{\Lambda_n} + || \ \gamma_{n} - s_{n} \ ||_{\Xi_n} \bigg],
 \label{eq:switch_con}
\end{equation}

\noindent where $S$ is the set of estimated measurement weights, $\psi()$ is a real-valued function\footnote{Within~\cite{sunderhauf2012robust} it is recommended to utilize a piece-wise linear function.} such that $\psi(*) \rightarrow [0,1]$, and $\gamma$ is a prior on the switch constraint.

To explicitly view the switch constraint technique as a covariance adaptation method, we can evaluate the effect that switchable constraints will have on measurement constraint in the modified cost function (e.g., Eq. \ref{eq:switch_con}), as depicted in Eq. \ref{eq:switch_cov_1} where $w_n \triangleq \psi(s_{n})$. This relation can be further reduced to the expression presented in Eq. \ref{eq:switch_cov_2}, where $\hat{\Lambda}_n^{-1} = w_n^2 \Lambda_n^{-1}$, by noting that $w_n$ is a scalar.

\begin{align}
 || \ \psi(s_{n}) r_n (X) \ ||_{\Lambda_{n}} & = \big[w_{n}r_{n}(X)\big]^T \Lambda^{-1}_{n} \big[w_{n}r_{n}(X)\big] \label{eq:switch_cov_1} \\&=r_n(X)^T \hat{\Lambda}_n^{-1} r_n(X)\label{eq:switch_cov_2}
\end{align}

The relation presented in Eq. \ref{eq:switch_cov_2} directly shows a relation between the estimated observation weighting --- as implemented in the switchable constraint framework --- and the scaling of the \textit{a priori} measurement covariance matrix. 

\subsubsection{Dynamic Covariance Scaling}

One potential disadvantage of the switchable constraint framework is the possibility for an increased convergence time due to the augmentation of the optimization space \cite{DCS}. To combat this issue, the \ac{DCS} \cite{DCS} approach was proposed as a closed-form solution to the switch constraint weighting optimization. The m-estimator equivalent weighting function, as derived within \cite{agarwal2015robust}, is provided as

\begin{equation}
 w\big( x \big)=\begin{cases}
 1, & \text{if $x^2 \leq k$}. \vspace{0.5em}\\
 \frac{4k^2}{(x^2 + k)^2}, & \text{otherwise},
 \end{cases}
 \label{eq:dcs_weight}
\end{equation}

\noindent where $k$ is the user defined kernel width, which dictates the residual threshold for observations to be considered erroneous. From this expression, it is noted that \ac{DCS} is equivalent to the standard least-squares estimator when the residuals are less than the kernel width $k$; however, when residuals lie outside of the kernel width, the corresponding observations are de-weighted according to the provided expression.

From the provided weighting expression, the equivalent \ac{DCS} robust m-estimator can be constructed, as presented in Table \ref{table:dcs_m-estimator}. This is performed by utilizing the definitions provided within \cite{zhang1997parameter}, (i.e.,  assuming a weighting function is provided, then the influence function is calculated as $\psi(x) = x \ w(x)$ and the robust cost function is $\rho(x) = \int \psi(x)dx$).

\begin{table}[htb!]
 \renewcommand{\arraystretch}{1.5}
 \centering
 \caption{Equivalent dynamic covariance scaling m-estimator, $w(x)$ is the weighting function -- as specified in Eq. \ref{eq:dcs_weight}, $\psi(X) \triangleq x\ w(x)$ is the influence function, $\rho(X) = \int \psi(x) dx$ is the robust cost function, and $k$ is the user defined kernel width.}

\begin{tabular}{|| c | c | c | c ||}
   \hline
   M-Estimator & $\rho(x)$ & $\psi(x)$ & $w(x)$ \\ [0.5ex]
   \hline\hline
   DCS $\left\{ \begin{array}{@{}ll@{}} \text{if} \ \ x^2 \leq k \\ \text{if} \ \ x^2 > k \end{array}\right.$
   & $\left\{ \begin{array}{@{}ll@{}} \frac{1}{2}x^2 \\ \frac{k\big(3x^2-k\big)}{2\big( x^2 +k \big)} \end{array}\right.$  &
   $\left\{ \begin{array}{@{}ll@{}} x \\  \frac{4k^2x}{(x^2 + k)^2}  \end{array}\right.$
   & $\left\{ \begin{array}{@{}ll@{}} 1 \\  \frac{4k^2}{(x^2 + k)^2} \end{array}\right.$ \\
   \hline
  \end{tabular}
 \label{table:dcs_m-estimator}
\end{table}

With the \ac{DCS} equivalent m-estimator, and the discussion provided in section \ref{subsubsec:m-est}, it can seen that \ac{DCS} based state estimation enables robustness through the adaptation of the {\textit{a priori}} measurement error covariance. The specific relation between the {\textit{a priori}} and modified covariance matrices is provided in Eq. \ref{eq:m-est_cov_scale}  with a  weighting function implementation as provided in Eq. \ref{eq:dcs_weight}.

\subsubsection{Max Mixtures}

While the previously discussed estimation frameworks do enable robust inference, they still have the undesirable property that the measurement uncertainty model must adhere to a uni-modal Gaussian model. To relax this assumption, a \ac{GMM} \cite{bishop2006pattern} can be utilized to represent the measurement uncertainty model. The \ac{GMM} characterizes the uncertainty model as a weighted sum of Gaussian components, as depicted in Eq. \ref{eq:gmm}, where \added{$r_n \triangleq y_n - h_n(X)$ is an observation residual, } \replaced{$w = \{w_1, w_2, \ldots, w_m\}$}{ $w$} is the set of mixture weights with the constraint that $\sum_{m=1}^M w_m = 1$, and $\mu_m$, $\Lambda_m$ are the mixture component mean and covariance, respectively.

\begin{equation}
 r_n \sim \sum_{m=1}^{M} w_m \mathcal{N}(r_n \ | \ \mu_m, \Lambda_m)
 \label{eq:gmm}
\end{equation}

Utilizing a \ac{GMM} representation of the measurement uncertainty model can greatly increase the accuracy of the measurement error characterization; however, it also greatly increases the complexity of the optimization problem. This increase in complexity is caused by the inability to reduce the factorization presented in Eq. \ref{eq:fg_optimization} to a \ac{NLLS} formulation, as was possible when a uni-modal Gaussian model was assumed.

To enable the utilization of a \ac{GMM} while minimizing the computational complexity of the optimization process, the max-mixtures approach was proposed within \cite{maxmix}. This approach circumvents the increased computational complexity by replacing the summation operation in the \ac{GMM} with the maximum operation, as depicted in Eq. \ref{eq:max_mix}, \added{where $\mathrel{\dot\sim}$ depicts that $r_n$ is approximately distributed according to the right-hand side}.

\begin{equation}
r_n \mathrel{\dot\sim} \max_m w_m \mathcal{N}(r_n \ | \ \mu_m, \Lambda_m)
 \label{eq:max_mix}
\end{equation}

 Within \replaced{Eq. \ref{eq:max_mix}}{ this formulation}, the maximum operator acts as a selector (i.e., for each observation, the maximum operator selects the single Gaussian component from the \ac{GMM} that maximizes the likelihood of the individual observation given the current state estimate). Through this process, each observation is only utilizing the single Gaussian component from the model, which allows for the simplification of the optimization process to the weighted sum of squared residuals \cite{dellaert2006square, dellaert2017factor} (i.e., a \ac{NLLS} optimization problem).

Similar to the previously discussed robust state estimation formulations, the max-mixtures implementation can also be interpreted as enabling robustness through covariance adaptation. Where, the \replaced{covariance adaptation}{ adaptation} is enabled through the maximum operator. Specifically, each observation initially belongs to a single component from the \ac{GMM} model, then through the iterative optimization process, the measurement error covariance for each observation is updated to one of the components from within the predefined \ac{GMM}.

\section{Proposed Approach}\label{sec:technical_approach}

As shown by the discussion in section \ref{sec:state_estimation}, several robust state estimation frameworks have been proposed. Each of \replaced{the discussed estimators}{ these methodologies} work under the assumption that an accurate {\textit{a priori}} measurement error covariance is available. However, in practice, the requirement to supply an accurate {\textit{a priori}} characterization of the measurement uncertainty model is not always feasible when considering that the platform could be operating in a novel environment\footnote{A novel environment is simply one that has not before been seen. \newline}, a non-cooperative environment\footnote{A non-cooperative environment is one that emits characteristics that inhibit the sensors ability to provide accurate observations.}, or both.

Within this section, \replaced{the \ac{BCE}}{ a novel} framework is proposed \replaced{to address}{ that addresses} the issue of robust state estimation where, the framework is not only robust to erroneous observations, but also erroneous {\textit{a priori}} measurement uncertainty estimates. To elaborate on the \replaced{\ac{BCE}}{ proposed} framework, the remainder of this section proceeds in the following manner: first, the assumed data model is discussed; then, a method for learning the measurement error uncertainty model from residuals is provided; finally, the \replaced{\ac{BCE}}{ proposed robust estimation} framework is described.

\subsection{Data Model} \label{subsec:data_model}

Given  a set of residuals $ R = \{r_1, r_2, \hdots, r_n\} $ with $r_n  = y_n - h_n(X) \in \mathbb{R}^d$. We will assume that the set can be partitioned into groupings of similar residuals (i.e., $\bigcup_{m=1}^{M} C_{m} = R$), where each group, $C_k$, can be fully characterized by a Gaussian distribution (i.e., $C_k \sim \mathcal{N}(\mu_k, \Lambda_k)$). Given this assumptions, the set of residuals are fully characterized as a weighted sum of Gaussian distributions (i.e., a \ac{GMM}), as depicted in Eq. \ref{eq:res_gmm}

\begin{equation}
 r_n \sim \sum_{m=1}^M w_m \mathcal{N}(r_n \ | \ \theta_m) \quad \text{s.t.} \quad \theta_m \triangleq \{ \mu_m, \Lambda_m\},
 \label{eq:res_gmm}
\end{equation}

\noindent where, $w$ is the set of mixture weights with the constraint that $ \sum_m w_m = 1$, and $\theta_m$, is the mixture components sufficient statistics (\replaced{i.e.,}{ i..e,} the mixture component mean and covariance).

To enable the explicit assignment of each data point to a Gaussian component within the \ac{GMM}, an additional latent parameter, \added{$Z = \{z_1, z_2, \ldots, z_n\}$ with $z_n \in \mathbb{R}^M$}, is incorporated. The assignment variables $Z$ are characterized according to a categorical distribution, as depicted in Eq. \ref{eq:catagorical}, where $w_m$ is the component weight, and $\operatorname{I}_{[*]}$ is the indicator function which evaluates to 1 if the equality within the brackets is true and 0 otherwise.

\begin{equation}
\operatorname{p}(z_n \ | \ w) = \prod_{m=1}^M w_m \operatorname{I}_{[z_n = m]}
 \label{eq:catagorical}
\end{equation}

Through the incorporation of the categorically distributed assignment parameters into the \ac{GMM}, every data instance $r_n \in R$ can be characterized through Eq. \ref{eq:full_mixture_model}. Where, this equation explicitly encodes that each data instance is characterized by a single Gaussian component from the full \ac{GMM}.

\begin{equation}
 \operatorname{p}(r_n \ | \ \theta_m, z_m ) = \sum_{m=1}^M w_m \mathcal{N}(r_n \ | \ \theta_m) \operatorname{I}_{[z_n = m]}
 \label{eq:full_mixture_model}
\end{equation}

As a Bayesian framework will be utilized for model fitting (as described in section \ref{subsec:gmm_fit}), a prior distribution must be defined for the \ac{GMM} mixture weights $w$, and the sufficient statistics, $\theta$. To define the prior over the mixture weights, $w$, a Dirichlet distribution \cite{frigyik2010introduction} is utilized, as it is a conjugate prior\footnote{Conjugate priors are commonly utilized because they make the calculation of posterior distribution tractable through the {\textit{a priori}} knowledge of the posterior distribution family \cite{diaconis1979conjugate}.\newline} to the categorical distribution \cite{tu2014dirichlet}. The specific form of the Dirichlet prior is provided in Eq. \ref{eq:dirichelt}, where $\alpha = (\alpha_1, \ldots, \alpha_{k})$\footnote{\added{For the analysis presented within this study, a symmetric Dirichlet prior was utilized (i.e., $\alpha_k = 1 \ \forall \ k$ ).}\newline} is a set of hyper-parameters such that $\alpha_i > 0$, and $S$ is the probability simplex\footnote{A simplex is simply a generalization of the triangle to n-dimensional space (i.e., $S = \lbrace w \in \mathbb{R}^m : w_i \geq 0 : \sum_m^M w_m = 1 \rbrace$ ).\newline}.

\begin{equation}
 \operatorname{p}(w | \alpha) = \frac{\Gamma(\sum_{m=1}^M(\alpha_M))}{\prod_{m=1}^M(\Gamma(\alpha_m))}\prod_{m=1}^{M}w_m^{\alpha_i-1} \operatorname{I}_{[w_m \in S]}
 \label{eq:dirichelt}
\end{equation}

\noindent To define a prior over the \ac{GMM} sufficient statistics, $\theta$, the normal inverse Wishart (NIW) is utilized due to the conjugate relation with the multivariate Gaussian that has unknown mean and covariance \cite{murphy2007conjugate}. The NIW distribution is defined in Eq. \ref{eq:wishart}, where $\beta = \{ m_o, \kappa_o, \nu_o, S_o\}$\footnote{\added{The NIW hyper-parameters can be intuitively understood as follows: $m_o$ is the expected prior on the mean vector $\mu$ with uncertainty $\kappa_o$, and $\nu_o$ is the expected prior on the covariance matrix $\Lambda$ with uncertainty $S_o$}} is the set of hyper-parameters, and $\operatorname{W}^{-1}$ is the inverse Wishart distribution, as discussed within \cite{alvarez2014bayesian}, where $\Gamma_d(*)$ is the multivariate gamma function, $\operatorname{Tr}(*)$ is the matrix trace, and $d$ is the dimension of $S_o$.

\begin{align}
 \operatorname{p}(\theta & \ |  \ \beta ) = \ \mathcal{N}(\mu | m_o, \frac{1}{\kappa_o}\Lambda) \operatorname{W}^{-1}(\Lambda| S_o,\nu_o)
 \label{eq:wishart}
 \\& \ \text{s.t.} \ \operatorname{W}^{-1}(\Lambda \ | \ S_o, \nu_o) = \frac{\lvert \Lambda^{ } \rvert^{ \frac{\nu_o - d - 1}{2} }}{ 2^{\nu_o \frac{d}{2}} \ {\Gamma_d \big( \frac{\nu_o}{2} \big) \ \lvert S_o \rvert^{ \frac{\nu_o}{2}}}  } e^{-\frac{1}{2}\operatorname{Tr}\big( S_o^{-1}\Lambda \big) }
 \nonumber
\end{align}

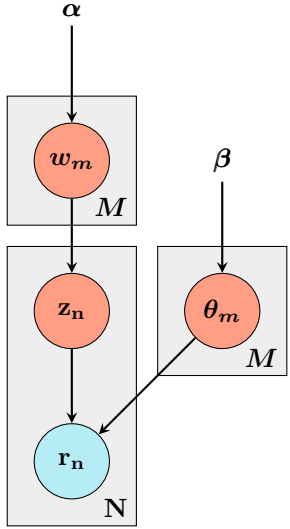
\begin{figure}
 \centering
 \begin{tikzpicture}[node distance=2cm]
  \node (alpha) {${\mathbf{\alpha}}$};
  \node (pi) [circ, below of=alpha] {$w_m$};
  \node (z) [circ, below of=pi] {$z_n$};
  \node (y) [circ, below of=z] {$r_n$};
  \node (beta) [right of=pi] {$\beta$};
  \node (theta) [circ, right of=z] {$\theta_m$};

  \node[inner sep=10pt,draw,fill=grey!20, fit=(pi) ] (piBox) {};
  \node[inner sep=10pt,draw,fill=grey!20,fit=(z) (y)] (zyBox) {};
  \node[inner sep=10pt,draw,fill=grey!20,fit=(theta)] (thetaBox) {};

  \node[above left] at (piBox.south east) {$\boldsymbol{M}$};
  \node[above left] at (zyBox.south east) {$\mathbf{N}$};
  \node[above left] at (thetaBox.south east) {$\boldsymbol{M}$};

  \node (alpha) {${\boldsymbol{\alpha}}$};
  \node (pi) [circ_open, below of=alpha] {$\boldsymbol{w_m}$};
  \node (z) [circ_open, below of=pi] {$\mathbf{z_n}$};
  \node (y) [circ, below of=z] {$\mathbf{r_n}$};
  \node (beta) [right of=pi] {$\boldsymbol{\beta}$};
  \node (theta) [circ_open, right of=z] {$\boldsymbol{\theta_m}$};

  \draw [arrow] (alpha) -- (pi);
  \draw [arrow] (pi) -- (z);
  \draw [arrow] (z) -- (y);
  \draw [arrow] (beta) -- (theta);
  \draw [arrow] (theta) -- (y);

 \end{tikzpicture}
 \caption{Graphical representation of the utilized data model which visually encodes the joint probability distribution presented in Eq. \ref{eq:joint_propb}. Each red node represents a latent parameter, and the blue node represents the provided observations. The plate notation \cite{jordan2004graphical} is utilized to represent the replication of a random variable.}
 \label{fig:dirProcess}
\end{figure}

Utilizing the data model defined earlier, the joint probability distribution that characterizes the system is defined in Eq. \ref{eq:joint_propb}. Additionally, a visual representation of the underlying model is graphically represented in Fig. \ref{fig:dirProcess}.

\begin{align}
 \operatorname{p}(R,\theta,Z) = & \operatorname{p}(w \ | \ \alpha) \cdot \nonumber \\&\prod_{n=1}^N \operatorname{p}(z_n \ | \ w_n) \operatorname{p}(r_n \ | \ z_n, \theta)\cdot \nonumber\\&\prod_{m=1}^M \operatorname{p}(\theta_m \ | \ \beta)
 \label{eq:joint_propb}
\end{align}

\subsection{Gaussian Mixture Model Fitting} \label{subsec:gmm_fit}

Provided with the joint distribution, as depicted in Eq. \ref{eq:joint_propb}, the ultimate goal of a \ac{GMM} fitting algorithm is to estimate the model parameters that maximize the $\operatorname{log \ marginal \ likelihood}$, as provided in Eq. \ref{eq:model_fit}, of the observations \cite{bishop2006pattern, steinberg2013unsupervised}. Due to the dimensionality of the space, the integral presented in Eq. \ref{eq:model_fit} is not computational tractable in general \cite{beal2003variational}. Thus, techniques for approximating the integral need to be utilized.

\begin{equation}
 \operatorname{log}\operatorname{p}(R) = \operatorname{log} \int \operatorname{p}(R, \theta, Z) d\mathbf{Z} d\boldsymbol{\theta}
 \label{eq:model_fit}
\end{equation}

In practice, two broad classes of integral approximation algorithms are utilized: Monte Carlo methods \cite{doucet2005monte}, and variational methods \cite{blei2017variational}. The Monte Carlo methods approximate the integral by averaging the response of the integrand for a finite number of samples\footnote{Considerable research is dedicated to efficient and intelligent ways to sample a space. For a succinct review of Monte Carlo sampling approaches, the reader is directed to \cite{doucet2005monte}.\newline}. In contrast, the variational approaches convert from an integration into an optimization problem, by defining a family of simplified functions (i.e., the family of exponential probability distributions) and optimizing the function within that family that most closely matches\footnote{Closeness in this context is usually measured with the Kullback-Leibler divergence \cite{beal2003variational, blei2017variational} or free-energy \cite{steinberg2013unsupervised}. \newline} the original function \cite{blei2017variational}.

Within this work, it was elected to utilize a variational model fitting approach, as this class of algorithms dramatically reduce the computational complexity of the problem when compared to Monte Carlo techniques \cite{beal2003variational}. To implement this approach, for the proposed data model, the assumption is made that the joint distribution defined in Eq. \ref{eq:joint_propb} can be represented as, $\operatorname{p}(R,\theta, Z) \approx \operatorname{q}(\theta)\operatorname{q}(Z)$, which is commonly referred to as the mean-field approximation \cite{blei2017variational}. Utilizing the mean-field approximation, the $\operatorname{log \ marginal \ likelihood}$ is represented as Eq. \ref{eq:mean_field}. Through the application of Jensens's inequality \cite{liao2018sharpening}, the equality in Eq. \ref{eq:mean_field} can be converted into a lower bound on the $\operatorname{log \ marginal \ likelihood}$\footnote{For a thorough review on the lower bound maximization for mixture model fitting, the reader is referred to \cite{harpaz2006algorithm}.}, as presented in Eq. \ref{eq:var_inf}.

\begin{align}
 \operatorname{log} \operatorname{p}(R) & = \operatorname{log} \int \operatorname{q}(Z) \operatorname{q}(\theta) \frac{\operatorname{p}(R,\theta, Z)}{\operatorname{q}(Z)\operatorname{q}(\theta)}dZd\theta \label{eq:mean_field} \\ &
 \begin{aligned}
 \ \ \geq                               & \int \operatorname{q}(Z)\operatorname{q}(\theta) \operatorname{log}\frac{p(R,Z \ | \ \theta)}{\operatorname{q}(Z)}dZd\theta \ +                                                         \\ &\int \operatorname{q}(\theta) \operatorname{log} \frac{\operatorname{p}(\theta)}{\operatorname{q}(\theta)} d\theta \label{eq:var_inf}
 \end{aligned}
\end{align}

The right hand side of the inequality presented in \replaced{Eq.}{ Eq.,} \ref{eq:var_inf} is commonly referred as the free energy functional \cite{steinberg2013unsupervised}, as depicted in Eq. \ref{eq:free_energy}. As it is a lower bound on the $\operatorname{log \ marginal \ likelihood}$, an optimal set of model parameters can be found through iterative optimization.

\begin{equation}
 \operatorname{log} \operatorname{p}(R) \geq \operatorname{F}\big[ \operatorname{q}(Z), \operatorname{q}(\theta) \big]
 \label{eq:free_energy}
\end{equation}

To find the set of equations necessary for iterative parameter optimization, the partial derivative of the free energy function can be taken with respect to $\operatorname{q}(Z)$, and $\operatorname{q}(\theta)$. When the partial derivative of the free energy functional is taken with respect to $\operatorname{q}(Z)$, as presented in Eq. \ref{eq:vbe_max}, the latent parameters, $Z$, can be updated by holding the model parameters, $\theta$, fixed and optimizing for $Z$. Likewise, when the partial derivative of the free energy functional is taken with respect to $\operatorname{q}(\theta)$, as presented in Eq. \ref{eq:vbe_exp}, the model parameters, $\theta$, can be updated by holding the latent parameters, $Z$, fixed and optimizing for $\theta$. Within Eq. \ref{eq:vbe_max} and Eq. \ref{eq:vbe_exp} the terms $C_z$ and $C_{\theta}$ are normalizing constants for $\operatorname{q}(Z)$ and $\operatorname{q}(\theta)$, respectively.

\begin{equation}
 \operatorname{q}(z)_{t+1} = C_z \int \operatorname{q}(\theta)_t \operatorname{log} p(R,Z \ | \ \theta) d\theta
 \label{eq:vbe_max}
\end{equation}

\begin{equation}
 \operatorname{q}(\theta)_{t+1} = C_\theta \operatorname{p}(\theta) \int \operatorname{q}(Z)_{t+1} \operatorname{log} \operatorname{p}(R,Z \ | \ \theta) dZ
 \label{eq:vbe_exp}
\end{equation}

When the variantional inference framework, as summarized in Eq. \ref{eq:vbe_max} and Eq. \ref{eq:vbe_exp}, is provided with a data set, $R$, which conforms to the models discussed in section \ref{subsec:data_model}, the output is a \ac{GMM} which characterizes the underlying probability distribution. Within this paper, the \ac{GMM} will be utilized to provide an adaptive characterization of the measurement uncertainty model.

\subsection{Algorithm Overview}

Multiple robust state estimation frameworks have been developed, as discussed in section \ref{sec:state_estimation}. However, it was noted that all of the discussed approaches fall short on at least one front. The primary shortcoming shared by all the approaches is the  inability to provide \replaced{an}{ a} accurate state estimate when provided with an inaccurate {\textit{a priori}} measurement error uncertainty model. To confront this concern, a novel \ac{IRLS} formulation is proposed within this section.

\begin{figure}
 \centering
 \begin{tikzpicture}[node distance=1.5cm]
  \node (start) [startstop, scale=0.75] {Construct Factor Graph};
  \node (pro1) [process, below of=start, scale=0.75] {\ac{NLLS} Optimization};
  \node (pro2) [process, below of=pro1, yshift=-1.0cm, scale=0.75] {Calculate Residuals};
  \node (pro3) [process, below of=pro2, scale=0.75] {Estimate Measurement Error Model};
  \node (pro4) [process, below of=pro3, scale=0.75] {Update Factor Graph Noise Model};
  \node (pro5) [process, below of=pro4, scale=0.75] {\ac{NLLS} Optimization};
  \node (dec1) [decision, below of=pro5, yshift=-1.2cm,scale=0.75] {Convergence};
  \node (stop) [startstop, below of=dec1, yshift=-1.0cm,scale=0.75] {Write Results};
  \draw [arrow] (start) -- (pro1);
  \draw [arrow] (pro1) -- (pro2);
  \draw [arrow] (pro2) -- (pro3);
  \draw [arrow] (pro3) -- (pro4);
  \draw [arrow] (pro4) -- (pro5);
  \draw [arrow] (pro5) -- (dec1);
  \draw [arrow] (dec1.west) node[above left, xshift=-0.7cm, scale=0.9]   {\large{\textbf{No}}} -- + (-25mm,0) |- (pro2.west);
  \draw [arrow] (dec1) -- node[anchor=east, scale=0.9] {\large{\textbf{Yes}}} (stop);

  \node (db1) [draw,inner sep=12pt, dashed, fit=(start) (pro1)] {};
  \node[above left] at (db1.north east) {\textbf{\small{Initialization}}};

  \node (db2) [draw,inner sep=12pt, dashed,fit=(pro2) (pro3) (pro4) (pro5)] {};
  \node[above left] at (db2.north east) { {\textbf{\small{Robust Iteration}} }};

 \end{tikzpicture}
 \caption{Schematical overview of the proposed \replaced{batch covariance estimation}{ robust optimized} algorithm. The proposed approach enables robust state estimation through the iterative estimation of the -- possible multimodal -- measurement error covariance model, where the measurement error covariance model is \replaced{characterized by a Gaussian mixture model that is fit to}{ estimated by clustering} the residuals of the previous iteration of optimization.}
 \label{fig:algo}
\end{figure}
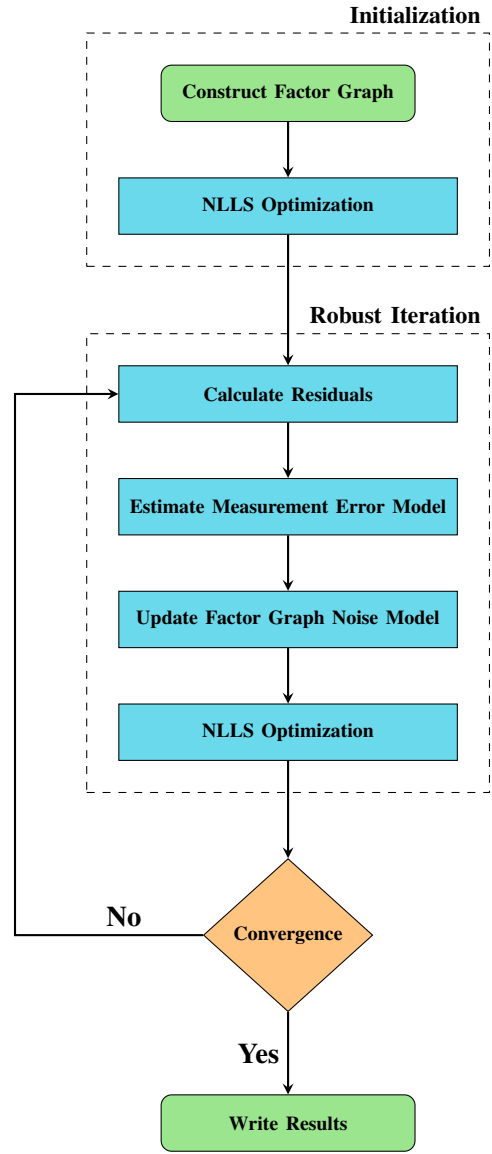

The proposed \added{\ac{BCE}} approach is graphically depicted in Fig. \ref{fig:algo}, where it is shown that the algorithm is composed of two sections: the initialization, and the robust iteration. To initialize the algorithm, a factor graph is constructed given the {\textit{a priori}} state and uncertainty information and the set of measurements. Utilizing the constructed factor graph, an initial iteration of the \ac{NLLS} optimizer is used to update the state estimate and calculate a set of measurement residuals.

With the residuals from the initial iteration of the \ac{NLLS} optimizer, the algorithm proceeds into the first robust iteration step. Each robust iteration step commences with the generation of a \ac{GMM} that characterizes the measurement uncertainty through the variational inference framework discussed in section \ref{subsec:gmm_fit}, operating on the most recently calculated set of residuals. With the provided \ac{GMM}, the measurement uncertainty model of the factor graph is updated (i.e., each measurement's uncertainty model is updated based upon the sufficient statistics of the assigned cluster from the \ac{GMM}). Finally, the state estimate is updated by feeding the modified factor graph into the \ac{NLLS} optimization algorithm. This robust iteration scheme is iterated until a measure of convergence\footnote{\added{To provide a measure of convergence, several criteria could be utilized. For this analysis the change in total error between consecutive iterations was utilized (i.e., the solution has converged to a minimum if the error between consecutive iterations is less than a predefined defined threshold).}\newline} -- or the maximum number of iterations\footnote{\added{The maximum number of iterations can be selected based upon a number of criteria (e.g., run-time consideration). For the analysis presented within this article, the maximum number of iterations was selected to equal 100.}} -- has been reached.

Because this framework iteratively updates a \ac{GMM} based measurement uncertainty model, the proposed approach is not only robust to erroneous measurements, but also robust to poor estimates of the {\textit{a priori}} measurement uncertainty model.

\section{Experimental Setup}\label{sec:experimental_setup}

\subsection{Data Collection}\label{subsec:data_collect}

To demonstrate the capabilities of our robust optimization framework, we evaluate our system using \ac{GPS} signals from a variety of environments, including open-air, and urban terrain. Due to the varying environments, the measurement uncertainty for each GPS measurements is unknown {\textit{a priori}} and differs over time.  Furthermore, using different qualities of GPS receivers also leads to differing measurement uncertainties.

\begin{figure*}[t!]
 \centering

 \begin{subfigure}[t]{0.33\textwidth}
  \centering
  \includegraphics[height=1.1in]{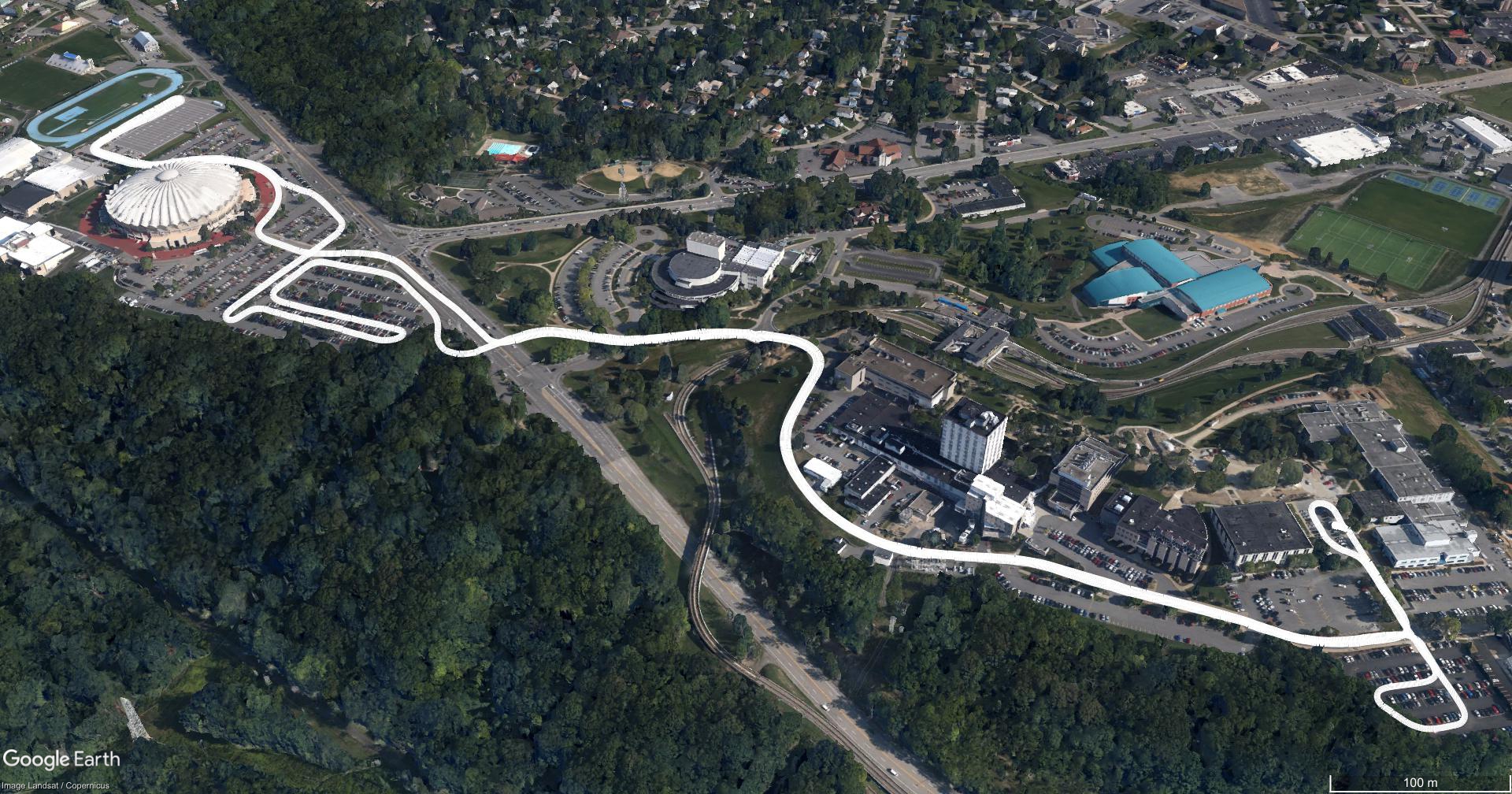}
  \caption{Ground trace for first data collect.}
 \end{subfigure}%
 ~
 \begin{subfigure}[t]{0.33\textwidth}
  \centering
  \includegraphics[height=1.1in]{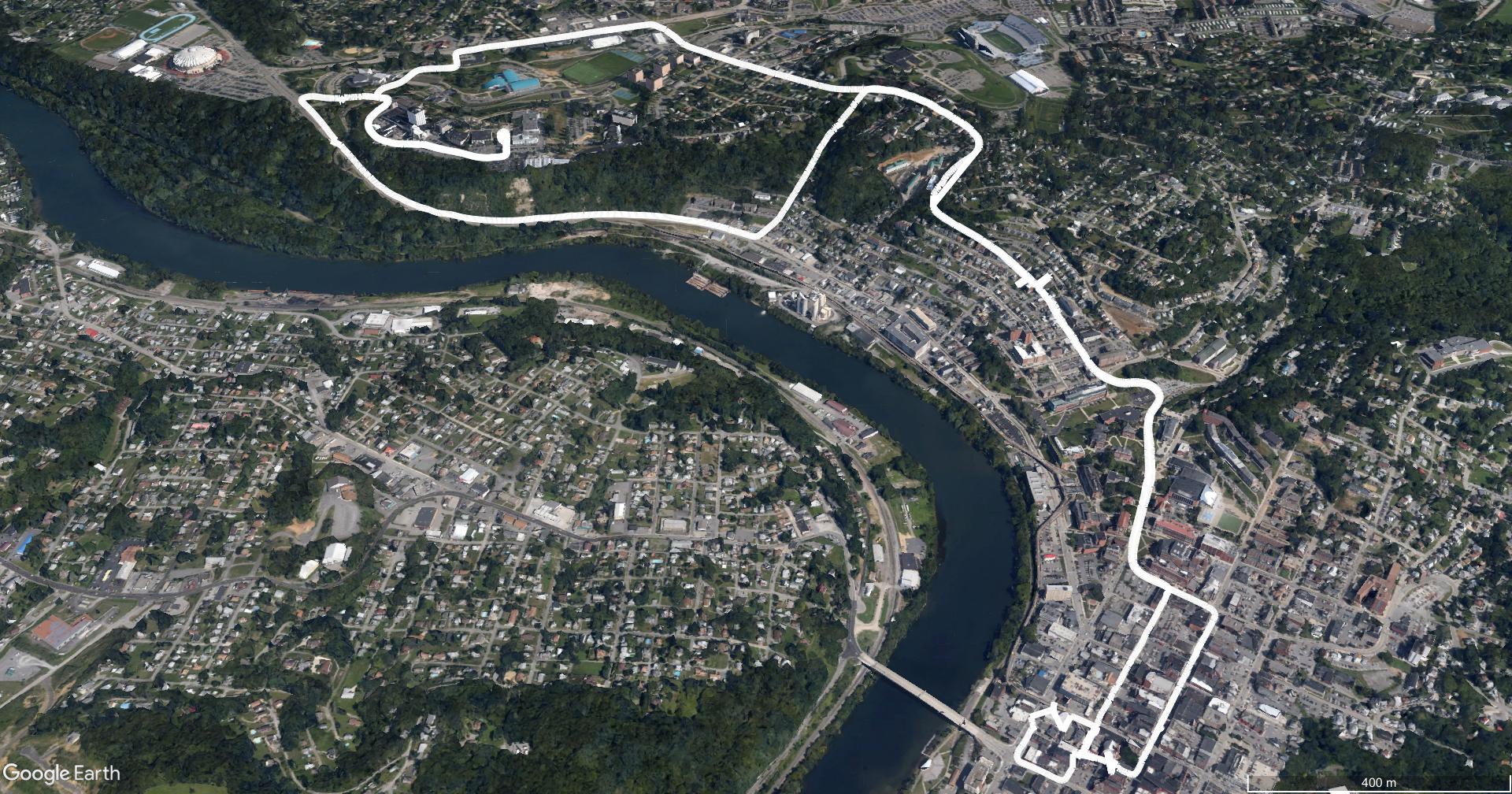}
  \caption{Ground trace for second data collect.}
 \end{subfigure}%
 ~
 \begin{subfigure}[t]{0.33\textwidth}
  \centering
  \includegraphics[height=1.1in]{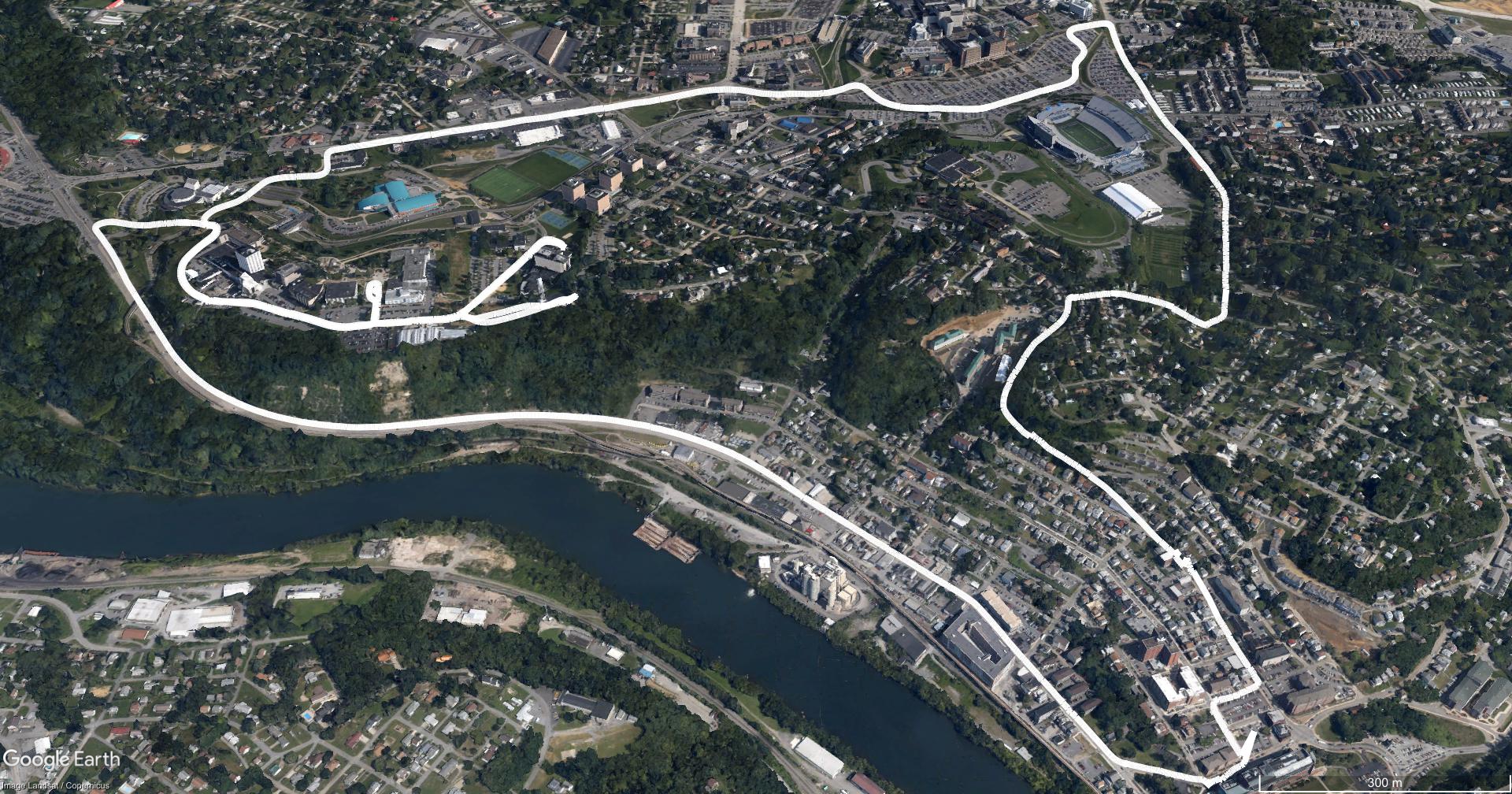}
  \caption{Ground trace for third data collect.}
 \end{subfigure}

 \caption{Ground trace of three kinematic driving data sets collected in the Morgantown, WV area.}
 \label{fig:ground_trace}

\end{figure*}

To test the proposed estimation approach using actual degraded \ac{GPS} observables, binary \ac{IQ} data in the L1-band was recorded using a LabSat-3 GPS record and playback device \cite{labSat3} during three kinematic driving tests, as depicted in Fig. \ref{fig:ground_trace}, in the Morgantown, WV area. These IQ data were then played back into both a geodetic-grade \ac{GNSS} receiver (Novatel OEM-638) and an open-source GPS \ac{SDR}, the GNSS-SDR library \cite{fernandez2001gnss}.  With this experimental setup, the geodetic-grade GNSS receiver was treated as a baseline reference for the best achievable GPS observable quality, and the GPS SDR was used to tune the quality of the observables to two different levels of performance ranging from low-grade to high-grade (i.e, close to matching the reference receiver). 





By altering the tracking parameters of the GPS SDR, it is possible to vary the accuracy of the GPS pseudorange and carrier-phase observables reported by the receiver.  In particular, by changing the tracking loop noise bandwidths and the spacing of the early and late correlators, both the level of thermal noise errors \cite{kaplan2005understanding} and the susceptibility to multipath errors can be varied \cite{van1992theory}. As detailed in \cite{kaplan2005understanding}, the thermal noise of the Phase Lock Loop (PLL) is directly proportional to the square root of the selected noise loop bandwidth, $B_{\phi}$

\begin{equation}
 \sigma_{PLL}= \frac{\lambda_{L1}}{2\pi}\sqrt{\frac{B_{\phi}}{C/N_0}\bigg(1+\frac{1}{2TC/N_0}\bigg)}
\end{equation}
where $\lambda_{L1}$ is the GPS L1 wavelength, $B_{\phi}$ is the carrier loop bandwidth in units of Hz, $C/N_0$ is the carrier-to-noise ratio in Hz, and $T$ is the integration time.

Likewise, the code tracking jitter of the \ac{DLL} is dependent on both the selected noise bandwidth, $B_{\rho}$, and the spacing of the early and late correlators, $D_{EL}$, however, this error model is also dependent on the bandwidth of the radio frequency frond-end, $B_{fe}$, of the receiver and the chiprate of the signal being tracked.  That is, if $D_{EL}$ is selected to be at or larger than $\pi$ multiplied by the ratio of the GPS C/A chipping rate, $R_{C/A}$, and  $B_{fe}$, then the approximation of the code tracking noise jitter (in units of chips) is given as \cite{kaplan2005understanding}
\begin{equation}
 \sigma_{DLL} = \sqrt{\frac{B_{\rho}}{2C/N_0}D_{EL}\bigg({1+\frac{2}{TC/N_0(2-D_{EL})}}\bigg)}
\end{equation}
if $\big( \frac{ R_{C/A}}{B_{fe}} < D_{EL} <  \frac{\pi R_{C/A}}{B_{fe}} \big)$ , then the code tracking jitter is approximated as \cite{kaplan2005understanding}
\begin{align*}
 \sigma_{DLL} = \Bigg[\frac{B_{\rho}}{2C/N_0}\Bigg(\frac{1}{B_{fe}}+\frac{B_{fe}T_c}{\pi-1}\bigg( D_{EL}-\frac{1}{B_{fe}T_c}\bigg)^2\Bigg) \\ \times \Bigg(1+\frac{2}{TC/N_0(2-D_{EL})}\Bigg)\Bigg]^{1/2}.
\end{align*}
Otherwise, if $\big( D_{EL} < \frac{R_{C/A}}{B_{fe}}\big)$, there is no additional benefit of further reducing $D_{EL}$, and the code noise error model reduces to a dependence on the front end and noise bandwidths\cite{kaplan2005understanding}.
\begin{equation}
 \sigma_{DLL}= \sqrt{\frac{B_{\rho}}{2C/N_0}\bigg(\frac{1}{B_{fe}T_c}\bigg)\bigg(1+\frac{1}{TC/N_0}\bigg)}
\end{equation}

For the experimental setup used in this study, the LabSat-3 has a $B_{fe}$ equal to 9.66 MHz and the GPS L1 C/A signal has a chiprate of 1.023 MHz, leading to a critical range of $D_{EL} \approx$ [0.11 .33].
An additional parameter that can be intuitively tuned in the GPS SDR configuration to provide an impact the observable quality is the rate at which the IQ data was sampled, $f_s$. In this setup, the LabSat-3 has a pre-defined sampling rate of 16.368 Mhz, but the data is down sampled to simulate reduced sampling quality in lower-cost receivers. Using these parameters, we are able to simulate a high-grade and low-grade quality GPS receiver using the parameters shown in Table \ref{table:SDRConfig}.

To quantify the accuracy of the generated observations, the zero-baseline test \cite{misra2006global} was implemented. A zero-baseline test, as its name suggests, performs a doubled-differenced differential GPS baseline estimation \replaced{between}{ bewtween} two sets of GPS receiver data that are known to have an a baseline of exactly zero.  That is, in this set-up, the non-zero estimated baseline magnitude is known to be due to discrepancies in the GPS observables reported by the two different receivers.  Using the high quality geodetic-grade GPS receiver observables as a reference, the zero-baseline tests provides a metric to assess the quality of the SDR receiver observables.

The zero-baseline test was implemented by first estimating the reference receiver solution and then configuring the RTKLIB DGPS software \cite{takasu2011rtklib} to estimate a moving baseline double-differenced GPS solution between the observables reported by the reference receiver and the observables reported by the particular GPS SDR tracking configuration. Because the observables are generated from the same RF front end (i.e., the LabSat-3) \added{and} connected to the same antenna, \added{the zero-baseline test results should provided zero difference, in an ideal scenario.} The \added{zero-baseline test results for the} two resulting GPS SDR cases, which are \replaced{utilized to experimentally validate the proposed approach within}{ used the experimental analysis in} this paper, are shown in Table \ref{table:zero_base_line}, where a large discrepancy can be see between the low-grade and high-grade observations.

\begin{table}
 \footnotesize
 \caption{GPS SDR observation tracking configurations utilized within this study.}
 \label{tab:lshape}
 \centering
  \begin{tabular}{||l|c|c|c|c||}
   \hline
   \vtop{\hbox{\strut GPS}\hbox{\strut Quality}} & $f_s$ (MHz) & $D_{EL}$ (chips) & $B_{\rho}$ (Hz) & $B_{\phi}$ (Hz) \\
   \hline\hline
   \rule{0pt}{2ex}Low                            & 4.092       & 0.5              & 2               & 50              \\
   \hline
   High                                          & 16.368      & 0.2              & 1               & 25              \\
   \hline
  \end{tabular}%
 \label{table:SDRConfig}
\end{table}

\begin{table}
 \centering
 \caption{Zero-baseline observation comparison for kinematic data sets -- see Fig. \ref{fig:ground_trace} -- with varied GPS SDR tracking configurations, as specified in Table \ref{table:SDRConfig}.}

  \begin{tabular}{||l|c|c|c||}
   \hline
   \vtop{\hbox{\strut Zero-Baseline Comparison}\hbox{\strut Mean 3D Error (m.)}} & Collect 1 & Collect 2 & Collect 3 \\
   \hline\hline
   \rule{0pt}{2ex}Low                                                            & 16.20     & 37.44     & 29.43     \\
   \hline
   High                                                                          & 0.59      & 18.03     & 17.48     \\
   \hline
  \end{tabular} %
 \label{table:zero_base_line}
\end{table}

The raw IQ recordings, GNSS-SDR processing configurations, and resulting observables in the \ac{RINEX} \cite{gurtner2007rinex}, that are used in this study have all been made publicly available at \href{https://bit.ly/2vybpgA}{https://bit.ly/2vybpgA}.

\subsection{GPS Factor Graph Model}

To utilize the collected GPS data within the factor graph framework, a likelihood factor must be constructed for the observations. To facilitate the construction of such a factor, the measurement model for the GPS L1-band pseudorange and carrier-phase observations will be discussed, as provided in Eq. \ref{eq:pseudorange} and Eq. \ref{eq:carrier_phase}, respectively. Within the pseudorange and carrier-phase observation models, the $X_s$ term is the known satellites location, $X_u$ is the estimated user location, and $||*||$ is the $l^2$-norm.

\begin{equation}
 \begin{split}
  \rho^{i}_{L_1}  = & \ ||X_s - X_u|| + c(\delta t^{}_{u} - \delta t_{s}) + T_{z,d}  \mathcal{M}_d(el^{j}) \\
  & \replaced{+}{ -} I_{L_1} + \delta_{Rel.} + \delta_{P.C.} + \delta_{D.C.B} + \epsilon^{j}_\rho
  \label{eq:pseudorange}
 \end{split}
\end{equation}

\begin{equation}
 \begin{split}
  \boldsymbol{\phi}^{i}_{L_1} = & \ ||X_s - X_u|| + c(\delta t^{}_{u} - \delta t_{s})+ T_{z,d} \mathcal{M}_{d}(el^{j}) \\
  & - I_{L_1} + \delta_{Rel.} + \delta_{P.C.} + \delta_{W.U.} + \lambda^{}_{IF} N^{j}_{IF} + \epsilon^{j}_\phi
  \label{eq:carrier_phase}
 \end{split}
\end{equation}

Contained within the models for the pseudorange and carrier-phase observations are several mutual terms, which can be decomposed into three categories. The first category is the propagation medium specific terms (i.e., the troposphere delay $T_{z,d}\mathcal{M}_d(el^j)$ \cite{niell1996global} and the ionospheric delay $I_{L1}$ \cite{klobuchar1987ionospheric}). The second category is the GPS receiver specific terms (i.e., the receiver clock bias $\delta t_u$, which must be incorporated into the state vector and estimated, and the observation uncertainty models $\epsilon$). The final category is the GPS satellite specific terms (i.e., the satellite receiver clock bias $\delta t_s$, the differential code bias correction $\delta_{D.C.B}$ \cite{kaplan2005understanding}, the relativistic satellite correction $\delta_{Rel.}$ \cite{pascual2007introducing}, and the satellite phase center offset correction $\delta_{P.C}$ \cite{heroux2001gps}).

Contained within the carrier-phase observation model (i.e., Eq. \ref{eq:carrier_phase}) are two terms that are not present in the pseudorange observation model. The first term is the carrier-phase windup \cite{wu1992effects}, which can easily be modeled to mitigate its effect. The second term is the carrier-phase ambiguity \cite{laurichesse2009integer}, which can not be easily modeled, and thus, must be incorporated into the state vector and estimated.

Utilizing the provided observation model, the GPS factor graph constraint can be constructed. To begin, we assume that the GPS observation uncertainty model is a uni-modal Gaussian. With this assumption, the GPS observations can be incorporated into the factor graph formulation using the mahalanobis distance \cite{sunderhauf2012multipath, watson2018evaluation}, as provided in Eq. \ref{eq:gnss_factor}, where $y$ is the observed GPS measurement, $\hat{y}$ is the modeled observation (calculated using Eq. \ref{eq:pseudorange} or Eq. \ref{eq:carrier_phase} for a pseudorange or carrier-phase observation, respectively), and $\Lambda$ is the measurement uncertainty model.

\begin{equation}
 \psi(i) = \rvert \rvert y - \hat{y} \lvert \lvert_{\Lambda}
 \label{eq:gnss_factor}
\end{equation}

With the GPS observation factor, the desired states can be estimated using any \ac{NLLS} algorithm \cite{sunderhauf2012multipath}, as discussed in section \ref{sec:state_estimation}. For this application, the state vector is defined in Eq. \ref{eq:state_vec}, where $\delta P$ is the 3-D user position state, $T_{z,w}$ is a state used to compensate for troposphere modeling errors, $\delta t_u$ is the receiver clock bias, and $N_*$ are the carrier-phase ambiguity terms for all observed satellites. For a thorough discussion on the resolution of carrier-phase ambiguity terms within the factor graph formulation, the reader is referred to \cite{watson2018evaluation}.

\begin{equation}
 \label{eq:state_vec}
 X = \begin{pmatrix}
 \delta P    \\
 T_{z,w}   \\
 \delta t_u \\
 N_1 \\
 \vdots \\
 N_n
 \end{pmatrix}
\end{equation}

To implement the provided model, this paper benefited from several open-source software packages. Specifically, to enable the implementation of the GPS observation modeling, the GPSTk software package \cite{harris2007gpstk} is utilized. For the factor graph construction and optimization, the Georgia Tech Smoothing and Mapping (GTSAM) library \cite{GTSAM} is leveraged.

\section{Results}\label{sec:results}

Utilizing the three kinematic data sets --- as depicted in Fig. \ref{fig:ground_trace} --- with varied receiver tracking metrics (i.e., the low quality and high quality receiver metrics, as discussed in Section \ref{sec:experimental_setup}), the proposed algorithm was tested with three state estimation techniques. The first algorithm used as a baseline comparison is a batch estimation strategy with an $l^2$-norm cost function. The second comparison algorithm is a batch estimator with the dynamic covariance scaling \cite{DCS} robust kernel. This specific algorithm was selected because it is both a closed-form solution to the switch constraints technique \cite{sunderhauf2012robust}, and a specific realization of the robust m-estimator. The final state estimation technique used to for this analysis is the max-mixtures \cite{maxmix} approach with a static measurement uncertainty model.

To generate the reference position solution which is utilized to enable a positioning error comparison, the GNSS observables reported by the reference GNSS receiver (i.e., a NovAtel OEM 638 Receiver) were used. Utilizing the baseline reference GPS measurements, the reference truth solution was generated through an iterative filter-smoother framework implemented within the open-source package RTKLIB \cite{takasu2011rtklib}.

\subsection{Positioning Performance Comparison}

\subsubsection{Low Quality Observations} \label{sec:low_quality}

\begin{figure}
 \centering
 \includegraphics[width=0.9\linewidth]{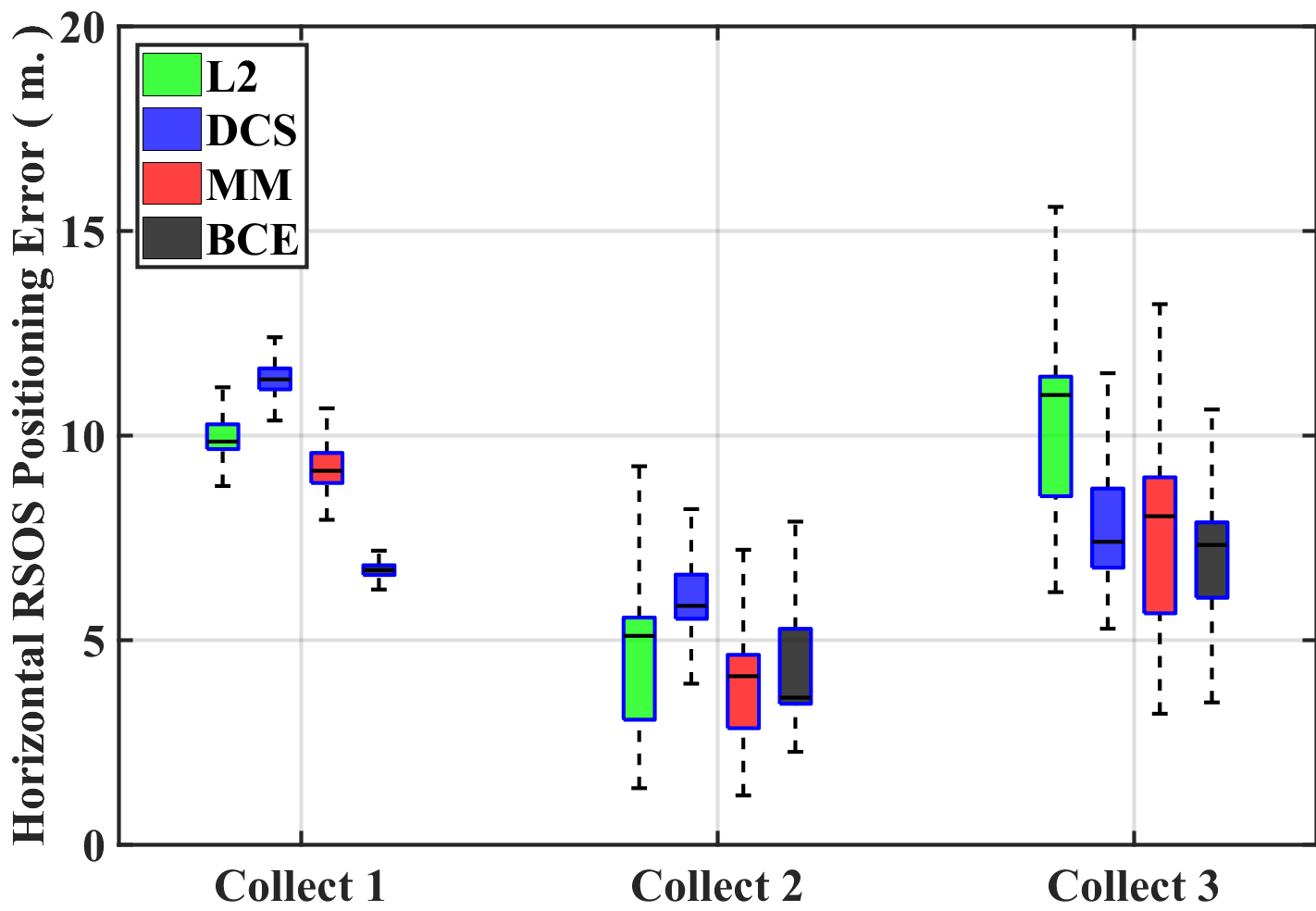}
 \caption{Box plot of RSOS positioning error for collected kinematic GNSS data sets with low quality -- see Table \ref{table:SDRConfig} for GNSS receiver configuration -- generated observations. The specific estimator statistics are provided in Table \ref{table:lq_stats}. Within this figure, $L2$, is a batch estimator with $l^2$ cost function, DCS is the dynamic covariance scaling robust estimator, MM is the max-mixtures approach with a static measurement covariance model, and BCE is the proposed batch covariance estimation technique. }
 \label{fig:lq_box}
\end{figure}

To begin an analysis, the discussed robust estimation techniques are evaluated when the algorithms are provided with low quality observations. As a metric to enable state estimation performance comparisons, the horizontal residual-sum-of-squares (RSOS) positioning error\footnote{The horizontal RSOS positioning error is utilized in place of the full 3-D RSOS positioning error due to the inability to accurately model the ionospheric delay with single frequency GPS observations, which imposes a constant bias for all estimations on the vertical positioning error.} is utilized. Utilizing the horizontal RSOS positioning metric, a solution comparison is provided in the form of a box-plot, as depicted in Fig. \ref{fig:lq_box}. From Fig. \ref{fig:lq_box}, it can be seen that the proposed algorithm significantly reduces the median horizontal positioning error when compared to the reference robust estimation techniques on the three collect data sets -- see Table \ref{table:lq_stats} for specific statistics. In addition to the median RSOS positioning error minimization, it is also noted that the proposed approach either outperforms or performs comparably to the three comparison approaches with respect to positioning solution variance and maximum error, as provided in Table \ref{table:lq_stats}.

\begin{table}
 \caption{Horizontal RSOS positioning error results when low quality observations are utilized. The green and red cell entries correspond to the minimum and maximum statistic, respectively.}
 \begin{subtable}{1.0\linewidth}
  \centering
  \caption{Horizontal RSOS positioning error results for data collect 1 when low quality -- see Table \ref{table:SDRConfig} for receiver configuration -- observations are utilized.}
  \begin{tabular}{||l|c|c|c|c||}
   \hline
   (m.)     & $L_2$ & DCS                       & MM                       & BCE                         \\
   \hline\hline

   median   & 9.84  & \cellcolor{red!60}{10.82} & 9.13                     & \cellcolor{green!60}{6.70}  \\
   \hline
   variance & 0.33  & 0.21                      & \cellcolor{red!60}{0.37} & \cellcolor{green!60}{0.09}  \\
   \hline
   max      & 14.84 & \cellcolor{red!60}{16.10} & 14.11                    & \cellcolor{green!60}{11.75} \\
   \hline
  \end{tabular}
  \label{table:drive_1_lq}
 \end{subtable}

 \begin{subtable}{1.0\linewidth}
  \centering

  \vspace{1em}
  \caption{Horizontal RSOS positioning error results for data collect 2 when low quality -- see Table \ref{table:SDRConfig} for receiver configuration -- observations are utilized.}
  \begin{tabular}{||l|c|c|c|c||}
   \hline
   (m.)     & $L_2$                    & DCS                          & MM                         & BCE                        \\
   \hline\hline

   median   & \cellcolor{red!60}{5.09} & 5.02                         & 4.11                       & \cellcolor{green!60}{3.58} \\
   \hline
   variance & 613.13                   & \cellcolor{green!60}{342.92} & \cellcolor{red!60}{673.50} & 393.32                     \\
   \hline
   max      & 127.29                   & \cellcolor{green!60}{98.05}  & \cellcolor{red!60}{132.49} & 103.64                     \\
   \hline
  \end{tabular}
  \label{table:drive_2_lq}
 \end{subtable}
 \begin{subtable}{1.0\linewidth}
  \centering
  \vspace{1em}
  \caption{Horizontal RSOS positioning error results for data collect 3 when low quality -- see Table \ref{table:SDRConfig} for receiver configuration -- observations are utilized.}
  \begin{tabular}{||l|c|c|c|c||}
   \hline
   (m.)     & $L_2$                     & DCS                        & MM                          & BCE                        \\
   \hline\hline

   median   & \cellcolor{red!60}{10.98} & 9.88                       & 8.02                        & \cellcolor{green!60}{7.31} \\
   \hline
   variance & 3.06                      & \cellcolor{green!60}{1.49} & \cellcolor{red!60}{3.36}    & 2.55                       \\
   \hline
   max      & 17.26                     & \cellcolor{red!60}{17.30}  & \cellcolor{green!60}{15.06} & 15.35                      \\
   \hline
  \end{tabular}
  \label{table:drive_3_lq}
 \end{subtable}
 \label{table:lq_stats}
\end{table}

\begin{figure}
 \centering
 \includegraphics[width=0.9\linewidth]{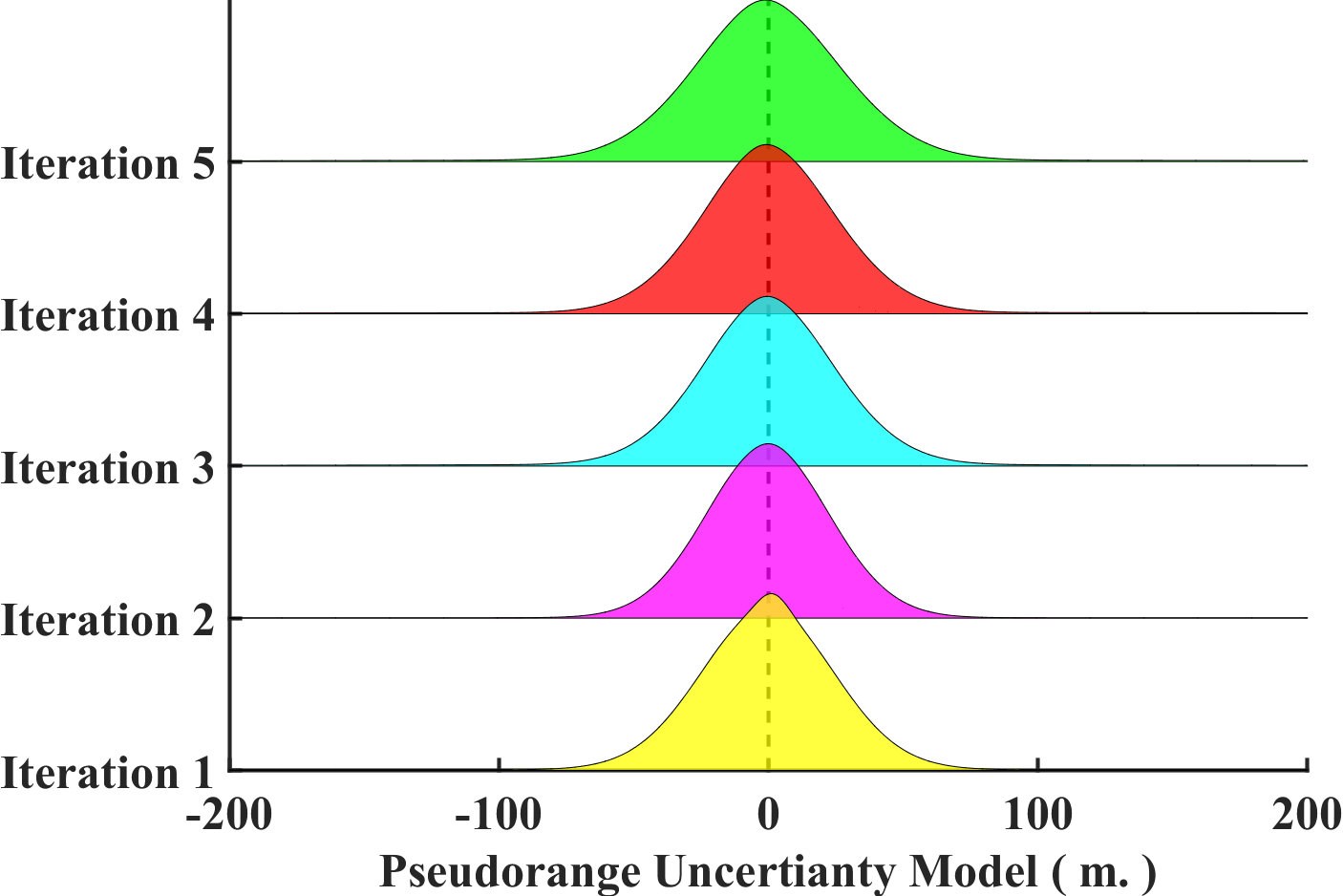}
 \caption{Evolution of the pseudorange Gaussian mixture model based uncertainty model for data collect 1 with low-quality observation as a function of optimizer iteration.}
 \label{fig:range_cov}
\end{figure}

\begin{figure}
 \centering
 \includegraphics[width=0.9\linewidth]{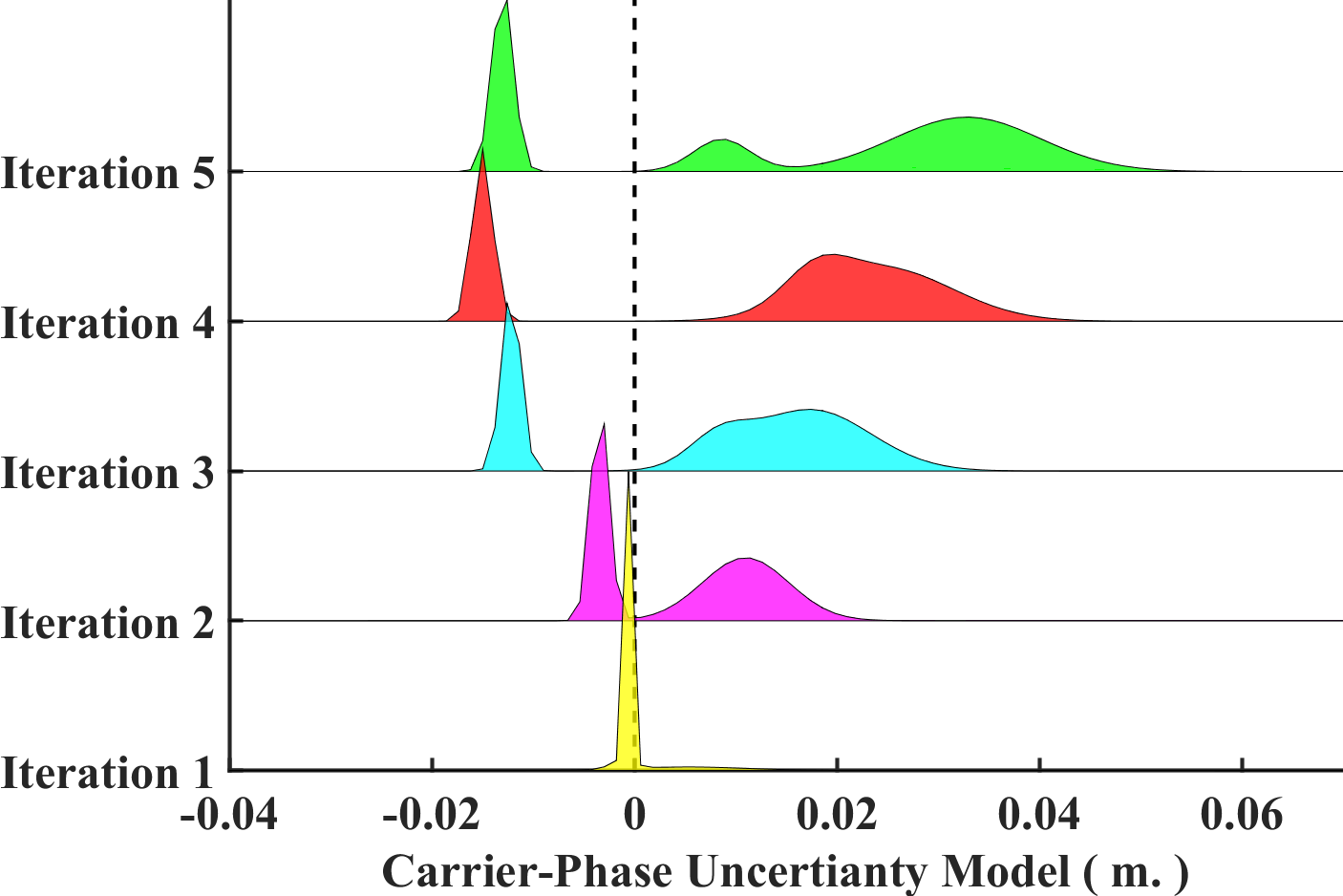}
 \caption{Evolution of the carrier-phase Gaussian mixture model based uncertainty model for data collect 1 with low-quality observation as a function of optimizer iteration.}
 \label{fig:phase_cov}
\end{figure}

\begin{figure*}[t!]
 \centering

 \begin{subfigure}[t]{0.3\textwidth}
  \centering
  \includegraphics[clip, width = \textwidth]{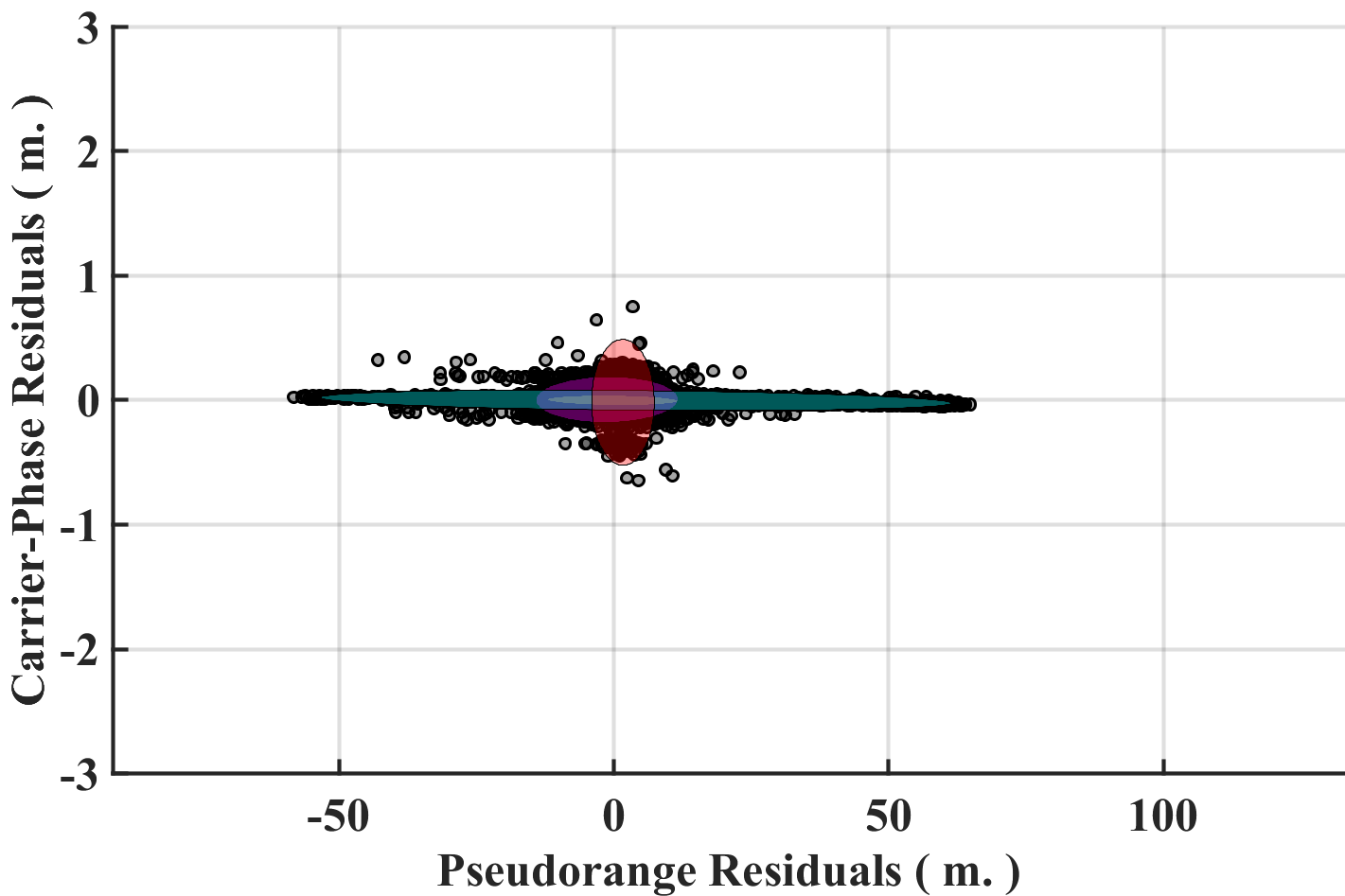}
  \caption{First Iteration}
  \label{fig:iter_0}
 \end{subfigure}%
 ~
 \begin{subfigure}[t]{0.3\textwidth}
  \centering
  \includegraphics[clip, width=\textwidth]{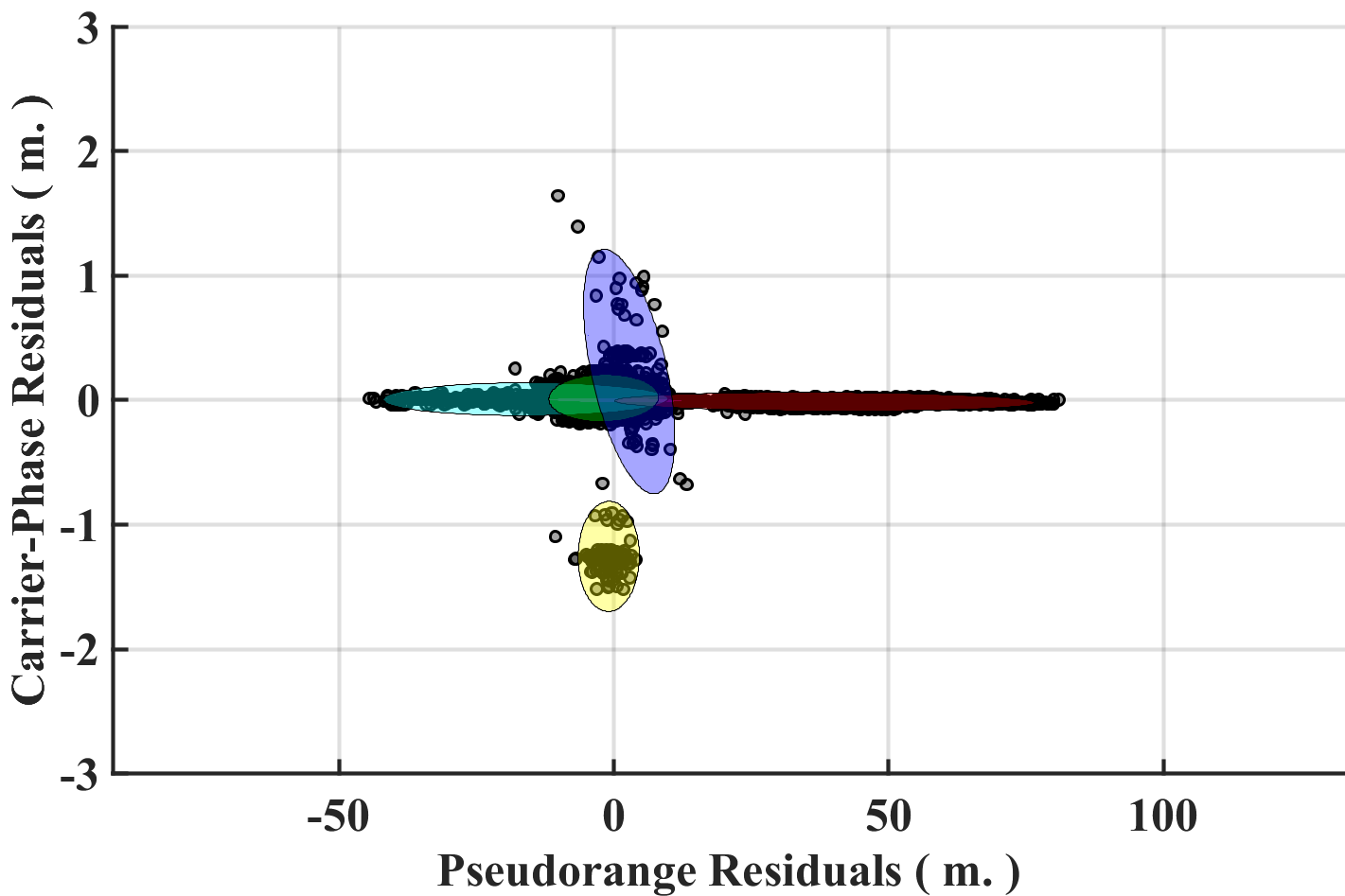}
  \caption{Third Iteration}
  \label{fig:iter_3}
 \end{subfigure}%
 ~
 \begin{subfigure}[t]{0.3\textwidth}
  \centering
  \includegraphics[clip, width=\textwidth]{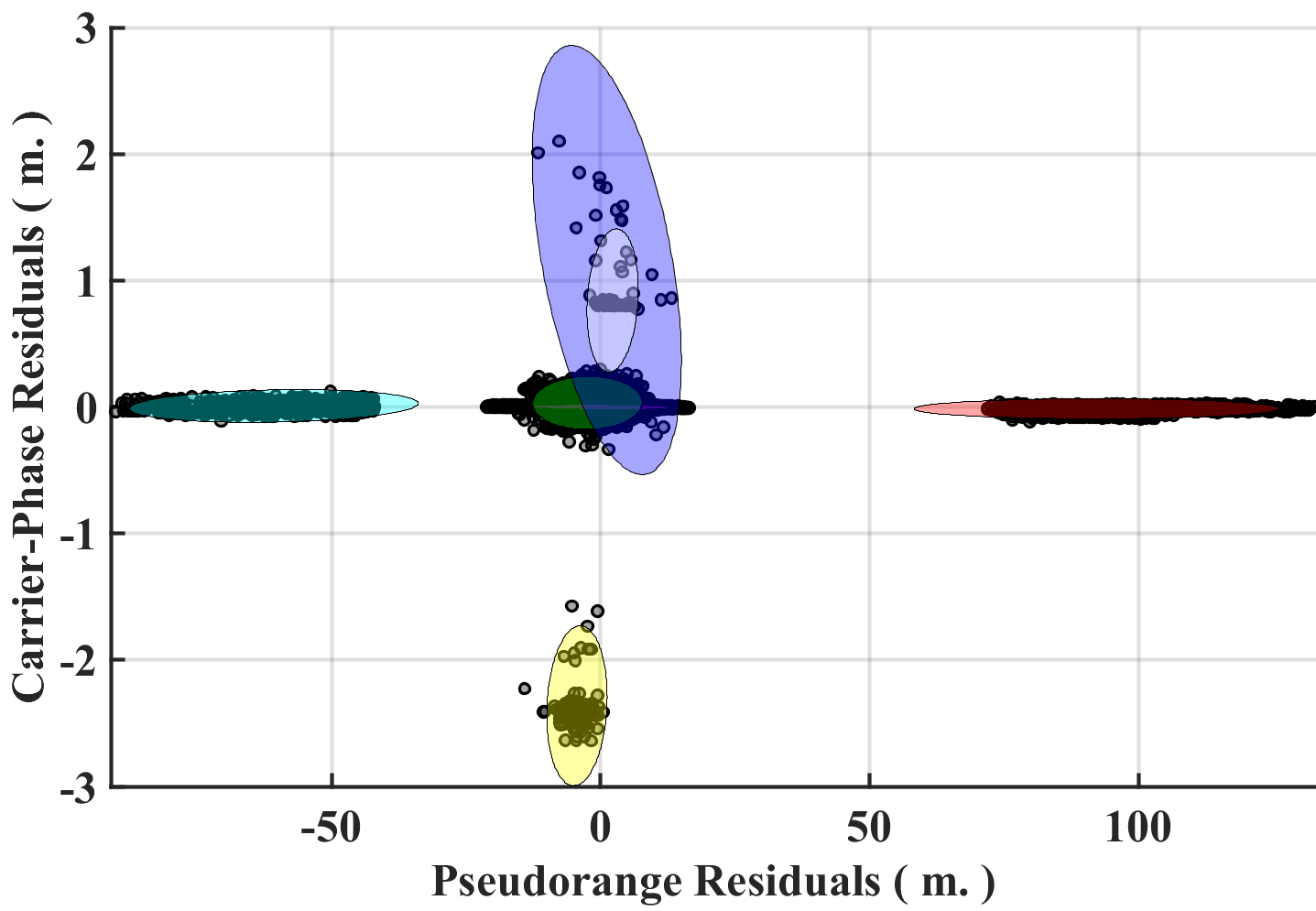}
  \caption{Fifth (Final) Iteration}
  \label{fig:iter_5}
 \end{subfigure}

 \caption{ Progression of GNSS observation residuals with multimodal measurement error covariance -- as estimated by the proposed approach which utilizes variational inference to generate a Gaussian mixture model uncertainty representation from the estimators residuals -- for data collect 1 with low-quality observations.}
 \label{fig:residual_progression}

\end{figure*}

To depict the adaptive nature of the measurement uncertainty model within the proposed approach, the estimated pseudorange and carrier-phase uncertainty models (for data collect 1 with low-quality observation) as a function of optimization iteration are depicted in Figures \ref{fig:range_cov}, and \ref{fig:phase_cov}, respectively. From Fig. \ref{fig:phase_cov} a large change in the structure of the assumed carrier-phase measurement uncertainty model can be seen. Specifically, the carrier-phase uncertainty model adapts from the assumed model (i.e, $y \sim \mathcal{N}(0,2.5\text{cm})$) to a highly multimodal \ac{GMM} at the final iteration, as depicted in Fig. \ref{fig:phase_cov}. As an additional visualization of the adaptive nature of the estimated covariance model, Fig. \ref{fig:residual_progression} is provided, which depicts the progression of the uncertainty model in the measurement residual domain for data collect 1 with low-quality observation as a function of the iteration of optimization.

\subsubsection{High Quality Observations} \label{sec:high_quality}

\begin{figure}
 \centering
 \includegraphics[width=0.9\linewidth]{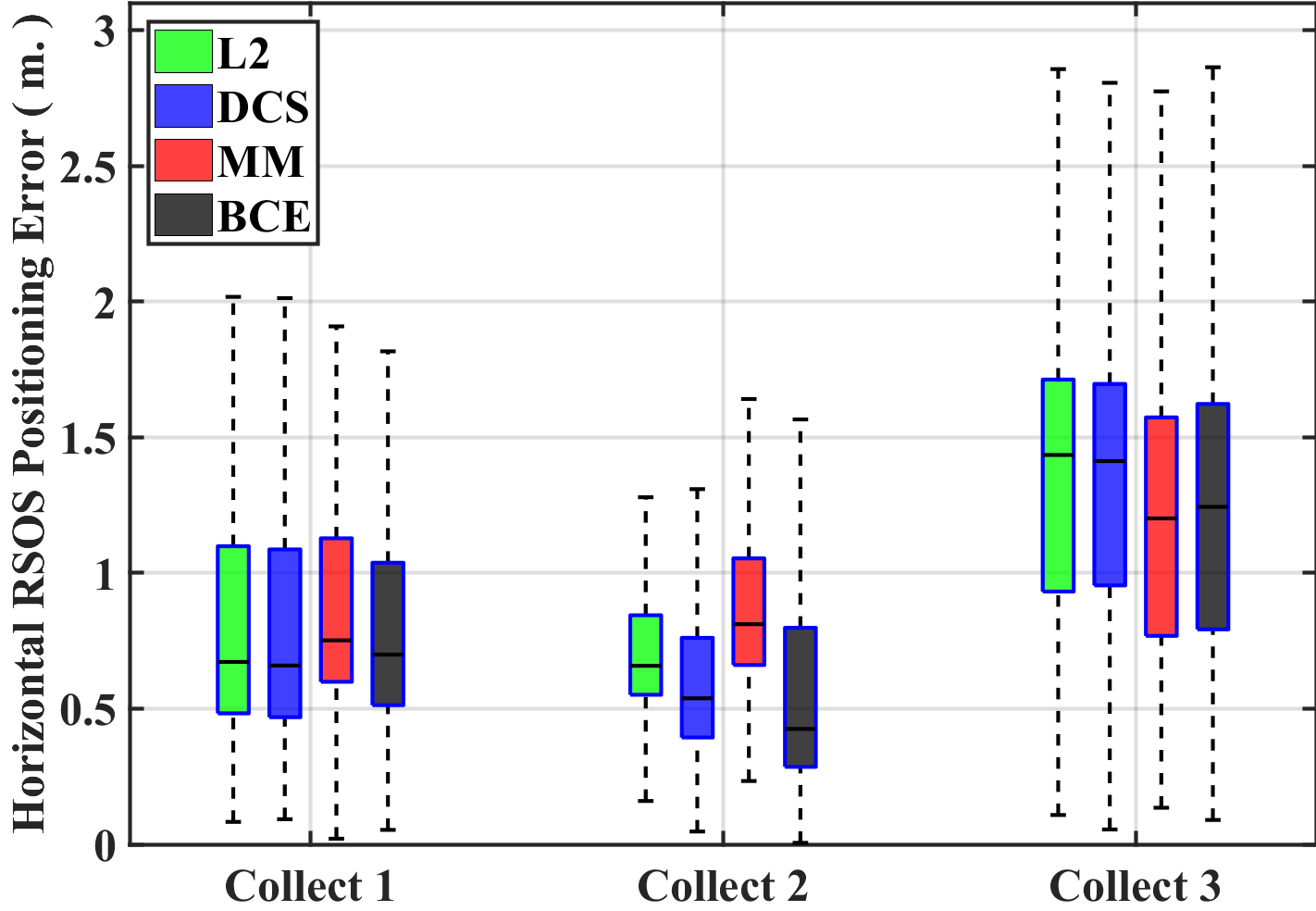}
 \caption{Box plot of RSOS positioning error for collected kinematic GNSS data sets with high quality -- see Table \ref{table:SDRConfig} for GNSS receiver configuration -- generated observations. The specific estimator statistics are provided in Table \ref{table:hq_stats}. Within this figure, $L2$, is a batch estimator with $l^2$ cost function, DCS is the dynamic covariance scaling robust estimator, MM is the max-mixtures approach with a static measurement covariance model, and BCE is the proposed batch covariance estimation technique. }
 \label{fig:hq_box}
\end{figure}

The high quality observations were also evaluated with the same four state estimation techniques. The horizontal RSOS positioning errors are depicted in box plot format in Figure \ref{fig:hq_box}. From Fig. \ref{fig:hq_box} it is shown that all four estimators provide similar horizontal positioning accuracy -- see Table \ref{table:hq_stats} for complete statistics.

The comparable horizontal RSOS positioning performance of all four estimation on the provided high quality observations is to be expected. This comparability, when provided with high quality observations, is because the {\textit{a priori}} assumed model closely resembles the observed model. This means that no benefit is granted by having a robust or adaptive state estimation framework in place.

\begin{table}
 \caption{Horizontal RSOS positioning error results when high quality observations are utilized. The green and red cell entries correspond to the minimum and maximum statistic, respectively.}
 \begin{subtable}{1.0\linewidth}
  \centering
  \caption{Horizontal RSOS positioning error results for data collect 1 when high quality -- see Table \ref{table:SDRConfig} for receiver configuration -- observations are utilized.}
  \begin{tabular}{||l|c|c|c|c||}
   \hline
   (m.)     & $L_2$                      & DCS                        & MM                       & BCE                        \\
   \hline\hline

   median   & 0.68                       & \cellcolor{green!60}{0.67} & \cellcolor{red!60}{0.75} & 0.69                       \\
   \hline
   variance & \cellcolor{red!60}{0.17}   & \cellcolor{red!60}{0.17}   & \cellcolor{red!60}{0.17} & \cellcolor{green!60}{0.15} \\
   \hline
   max      & \cellcolor{green!60}{6.30} & 6.30                       & \cellcolor{red!60}{6.37} & 6.34                       \\
   \hline
  \end{tabular}%
  \label{table:drive_1_hq}
 \end{subtable}
 \begin{subtable}{1.0\linewidth}
  \centering
  \vspace{1em}
  \caption{Horizontal RSOS positioning error results for data collect 2 when high quality -- see Table \ref{table:SDRConfig} for receiver configuration -- observations are utilized.}
  \begin{tabular}{||l|c|c|c|c||}
   \hline
   (m.)     & $L_2$                       & DCS                         & MM    & BCE                        \\
   \hline\hline

   median   & 0.65                        & 0.65                        & 0.81  & \cellcolor{green!60}{0.42} \\
   \hline
   variance & \cellcolor{green!60}{11.27} & \cellcolor{green!60}{11.27} & 11.94 & \cellcolor{red!60}{16.60}  \\
   \hline
   max      & \cellcolor{green!60}{18.29} & 18.30                       & 18.91 & \cellcolor{red!60}{21.73}  \\
   \hline
  \end{tabular}%
  \label{table:drive_2_hq}
 \end{subtable}
 \begin{subtable}{1.0\linewidth}
  \centering
  \vspace{1em}
  \caption{Horizontal RSOS positioning error results for data collect 3 when high quality -- see Table \ref{table:SDRConfig} for receiver configuration -- observations are utilized.}
  \begin{tabular}{||l|c|c|c|c||}
   \hline
   (m.)     & $L_2$                    & DCS                       & MM                         & BCE                        \\
   \hline\hline

   median   & \cellcolor{red!60}{1.44} & 1.43                      & \cellcolor{green!60}{1.20} & 1.24                       \\
   \hline
   variance & \cellcolor{red!60}{0.30} & \cellcolor{red!60}{0.30}  & \cellcolor{red!60}{0.30}   & \cellcolor{green!60}{0.26} \\
   \hline
   max      & 9.52                     & \cellcolor{red!60}{17.30} & \cellcolor{green!60}{9.43} & 9.44                       \\
   \hline
  \end{tabular}%
  \label{table:drive_3_hq}
 \end{subtable}
 \label{table:hq_stats}
\end{table}

\begin{figure}
 \centering
 \includegraphics[width=0.9\linewidth]{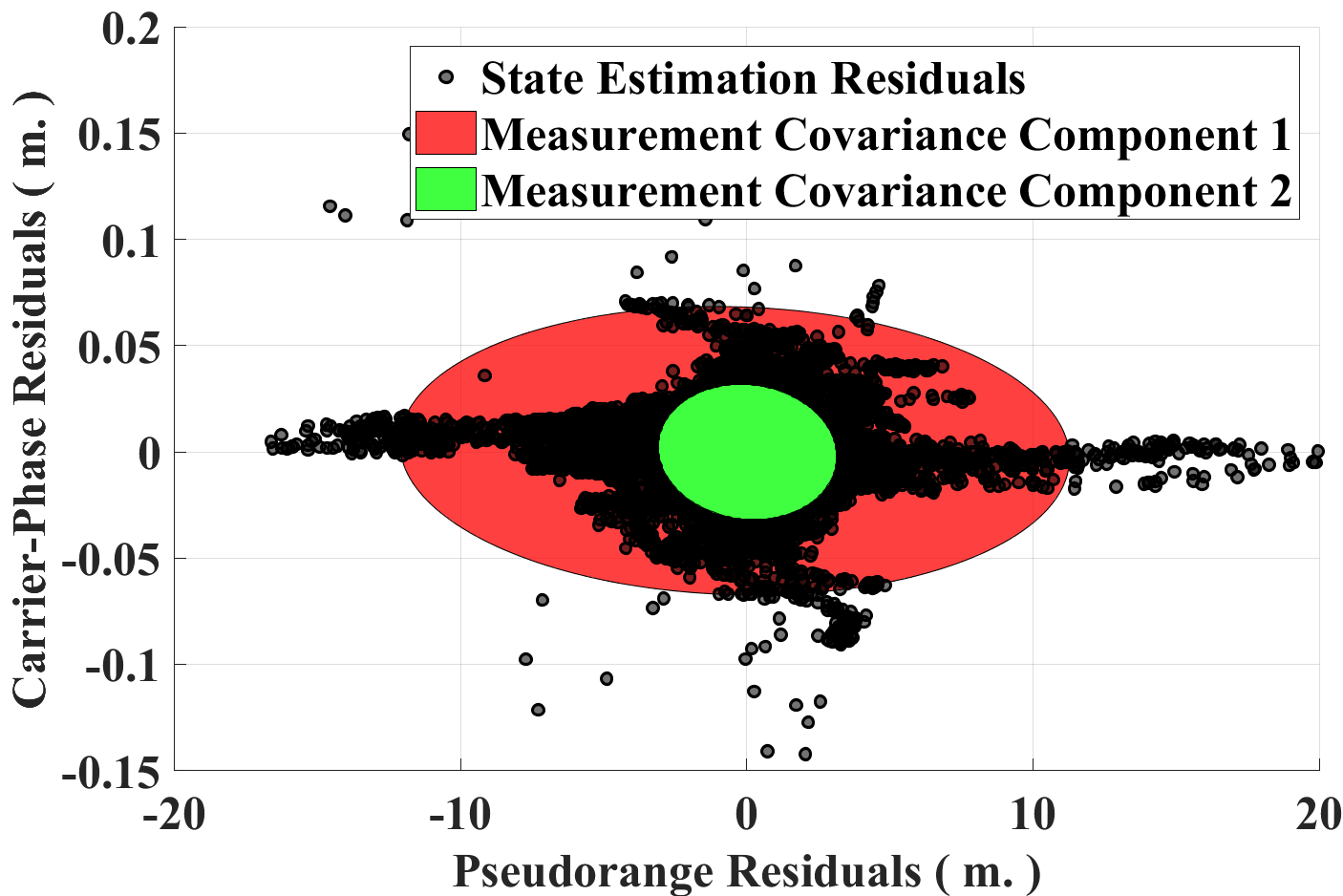}
 \caption{Measurement residuals with associated BCE estimated measurement error covariance -- each ellipse represents 95\% confidence -- for data collect 1 with high quality observations.}
 \label{fig:hq_residuals}
\end{figure}

To verify that the \replaced{assumed}{ assume} measurement model closely resembles the observed model, the estimated covariance of the proposed approach can be examined for a specific data set (i.e., data collect 1, as depicted in Fig. \ref{fig:hq_residuals}). For data collect 1 with high quality observations, the measurement uncertainty model estimated by the proposed approach has two modes, as depicted in Fig. \ref{fig:hq_residuals}. One of the modes --- specifically, the mode that characterizes approximately 90\% of the measurements --- closely resembles the assumed {\textit{a priori}} measurement uncertainty model. For comparison, the specific values of the {\textit{a priori}} measurement uncertainty model ($\Lambda$), and the estimated measurement uncertainty model ($\hat{\Lambda}$) as provided by the proposed approach, are provided below.

$$\Lambda =  \left[ {\begin{array}{*{20}c} 2.5^2 & 0.0  \\ 0.0 & 0.025^2  \\ \end{array} } \right] \quad \quad \hat{\Lambda} =  \left[ {\begin{array}{*{20}c} 1.7^2 & 0.0  \\ 0.0 & 0.0017^2  \\ \end{array} } \right] $$

\subsection{\textit{A Priori} Information Sensitivity Comparison}

\added{To continue the analysis, an evaluation of each estimation framework's sensitivity to the \apre information is provided. Specifically, it is of interest to evaluate the sensitivity of the estimation frameworks to the \apre measurement error covariance model.}

\added{To enable this estimator sensitivity evaluation, the low-quality observations generated from data collect 1 are utilized. With these generated observations, the estimation framework's sensitivity is quantified by evaluating the response -- which is quantified by the horizontal RSOS positioning error -- of each estimator as a function of the \apre covariance model. For this study, the \apre covariance model will be a scaled version of the assumed measurement error covariance model, as depicted below.}

$$\Lambda =   s \cdot \left[ {\begin{array}{*{20}c} 2.5^2 & 0.0  \\ 0.0 & 0.025^2  \\ \end{array} } \right]$$

\begin{figure}
 \centering
 \includegraphics[width=0.9\linewidth]{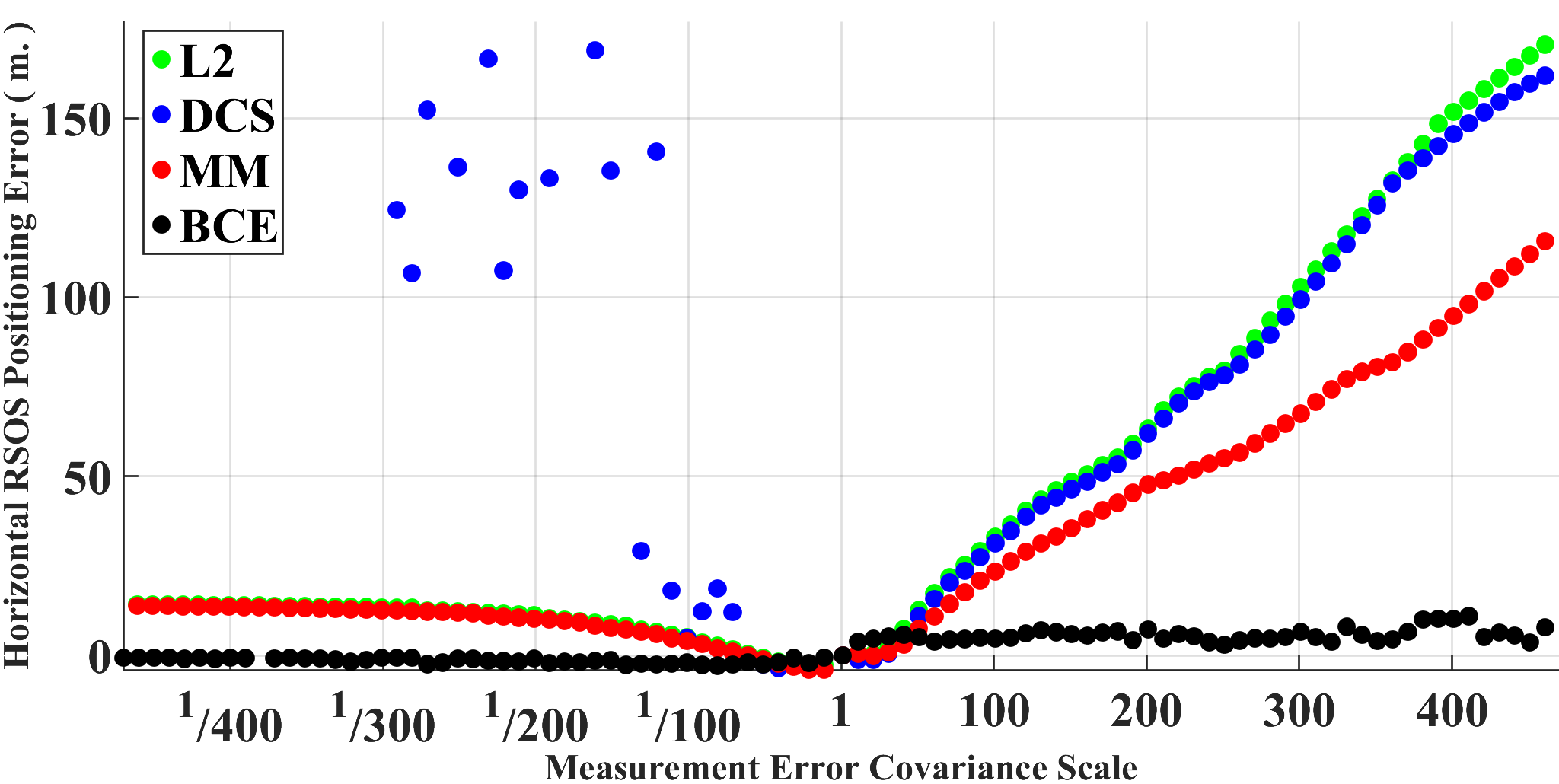}
 \caption{\added{Sensitivity of the estimation frameworks to the selection of an \apre measurement error covariance model. Within this figure, $L2$ is a batch estimator with $l^2$ cost function, DCS is the dynamic covariance scaling robust estimator, MM is the max-mixtures approach with a static measurement covariance model, and BCE is the proposed batch covariance estimation technique.}}
 \label{fig:cov_sensitivity}
\end{figure}

\added{This sensitivity evaluation is presented graphically within Fig. \ref{fig:cov_sensitivity}. From this figure, it is shown that all the estimation frameworks show at least a slight sensitivity to the \apre measurement error covariance model; however, some estimation frameworks are significantly more sensitive to perturbation in the \apre covariance model than other (e.g., the L2 and DCS estimators showing the greatest sensitivity). From this figure, it is also shown that the proposed \ac{BCE} approach is significantly less sensitivity to perturbations of the \apre measurement error covariance model when compared to the other estimators.}

\subsection{Run-time Comparison}

\added{To conclude the evaluation of the proposed estimation framework, a run-time comparison is provided. To quantify a run-time evaluation, the wall-clock time (i.e., the total execution time of the utilized estimation framework) divided by the cardinality of the utilized data collect is employed as the metric of comparison. This metric is selected because it enables a run-time comparison regardless of the data collect (i.e., each data collect has a different number of observations so, dividing by the cardinality enables a fair run-time comparison across sets). To implement the run-time comparison, a quad-core Intel i5-6400 central processing unit (CPU) that has a base processing frequency of 2.7 GHz was utilized.}

\begin{figure}
 \centering
 \includegraphics[width=0.9\linewidth]{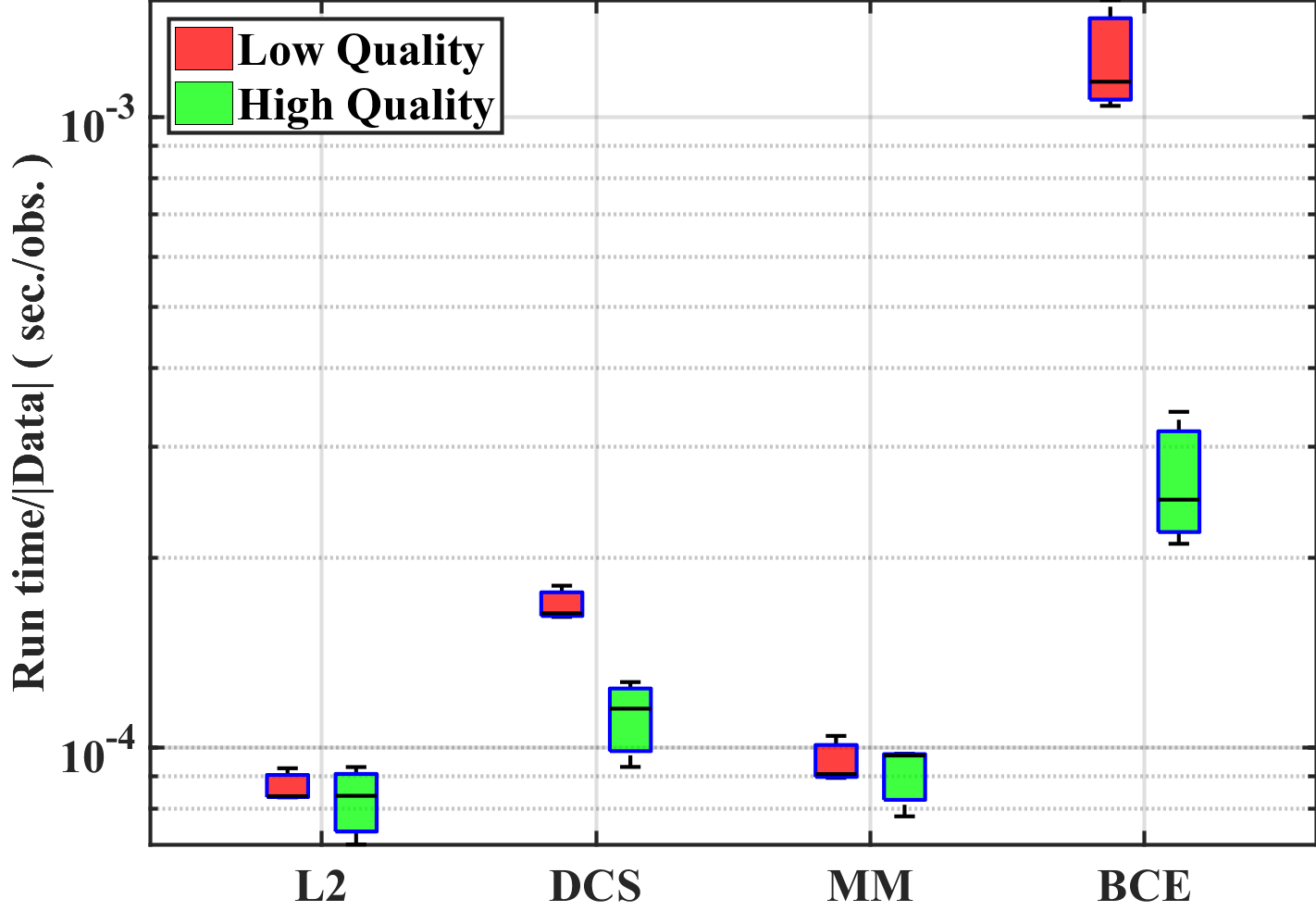}
 \caption{\added{Wall-clock time per utilized observation for each of the estimation frameworks, where $L2$ is a batch estimator with $l^2$ cost function, DCS is the dynamic covariance scaling robust estimator, MM is the max-mixtures approach with a static measurement covariance model, and BCE is the proposed batch covariance estimation technique.}}
 \label{fig:run_time}
\end{figure}

\added{Utilizing the specified hardware and evaluation metric, the run-time comparison is presented within Fig. \ref{fig:run_time} for the four estimation algorithms. From Fig. \ref{fig:run_time}, it is shown that the traditional $l^2$ approach provides the fastest run-time, with the max-mixtures approach providing comparable results. Additionally, it is shown that the batch covariance estimation approach is the most computationally expensive approach.}

\added{The increased computational complexity of the proposed approach is primarily due to the \ac{GMM} fitting procedure. To reduce the computation requirement of the proposed estimation framework, several research directions could be explored. For example, rather than utilizing all calculated residuals to characterize the measurement uncertainty model, a sub-sampling based approach per iteration could be utilized~\cite{rocke2003sampling}. This modification has the potential to significantly decease the run-time of the employed variational clustering algorithm which, in turn, would decrease the run-time of the associated \ac{BCE} approach.}

\section{Conclusion}\label{sec:conclusion}

Several robust state estimation frameworks have been proposed over the previous decades. Underpinning all of these robust frameworks is one dubious assumption. Specifically, the assumption that an accurate ${\textit{a priori}}$ measurement uncertainty model can be provided. As systems become more autonomous, this assumption becomes less valid (i.e., as systems start operating in novel environments, there is no guarantee that the assumed {\textit{a priori}} measurement uncertainty model characterizes the sensors current observation uncertainty).

In an attempt to relax this assumption, a novel robust state estimation framework is proposed. The proposed framework enables robust state estimation through the iterative adaptation of the measurement uncertainty model. The adaptation of the measurement uncertainty model is granted through non-parametric clustering of the estimators residuals, which enables the characterization of the measurement uncertainty via a Gaussian mixture model. This Gaussian mixture model based measurement uncertainty characterization can be incorporated into any non-linear least square optimization routine by only using the single assigned --- the assignment of each observation to a single mode within the mixture model is provided by the utilized non-parametric clustering algorithm --- component's sufficient statistics from within the mixture model to update the uncertainty model for all observation (i.e., every observations uncertainty model is approximately characterized by the single assigned Gaussian component from within the mixture model).

To verify the proposed algorithm, several GNSS data sets were collected. The collected data sets provide varying levels of observation degradation to enable to characterization of the proposed algorithm on a diverse data set. Utilizing these data sets, it is shown that the proposed technique exhibits improved state estimation accuracy when compared to  other robust estimation techniques when confronted with degraded data quality.
\section*{Acknowledgment}

This work was supported through a sub-contract with MacAualy-Brown Inc. 


\bibliography{content/batch_cov}
\bibliographystyle{IEEEtran}


\end{document}